%% file: neurips_2025.tex
\documentclass{article}

\PassOptionsToPackage{numbers, compress}{natbib}

\usepackage[final]{neurips_2025}

\usepackage[numbers]{natbib}
\usepackage[utf8]{inputenc} 
\usepackage[T1]{fontenc}    
\usepackage{hyperref}       
\usepackage{url}            
\usepackage{booktabs}       
\usepackage{amsfonts}       
\usepackage{nicefrac}       
\usepackage{microtype}      
\usepackage{xcolor}         
\usepackage{adjustbox}

\setcitestyle{square}
\setcitestyle{comma}
\usepackage{hyperref}
\hypersetup{
	colorlinks=true,
	citecolor=teal,
}
\usepackage{url}
\usepackage{booktabs}
\usepackage{microtype}      
\usepackage{multicol}
\usepackage{amsmath}
\usepackage{enumitem}
\usepackage{amssymb}
\usepackage{fontawesome5}
\numberwithin{equation}{section} 
\usepackage{wrapfig,lipsum,booktabs}
\usepackage{xcolor}
\usepackage{multirow}
\usepackage{float}
\usepackage{graphicx}
\usepackage{subcaption}
\usepackage{algpseudocode}  
\usepackage{listings}
\usepackage{algorithmicx,algorithm}
\usepackage{caption}
\usepackage{diagbox} 
\usepackage{colortbl}
\usepackage{xcolor}
\definecolor{rowcolor}{rgb}{0.9, 0.9, 0.9}
\usepackage{soul}
\usepackage{pifont}
\usepackage{enumitem}
\let\oldding\ding
\renewcommand{\ding}[2][1]{\scalebox{#1}{\oldding{#2}}}

\newcommand{\cmark}{\textcolor[rgb]{0,0.5,0}{\checkmark}} 
\usepackage{makecell}    

\algnewcommand{\LineComment}[1]{%
  \Statex \textbf{\#}\,\ttfamily #1\normalfont}
\definecolor{darkgreen}{RGB}{0,100,0} %

\title{\raisebox{-0.3em}{\includegraphics[width=0.7cm]{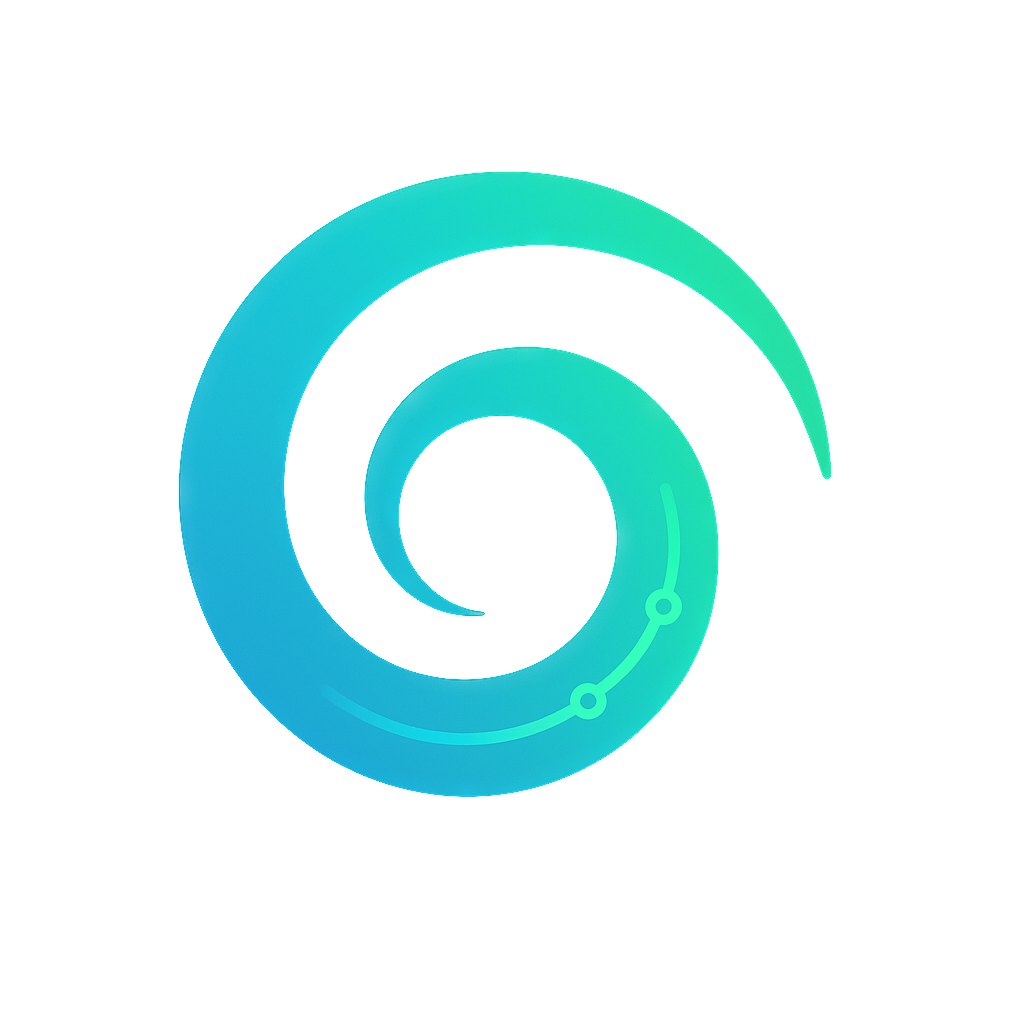}}InfantAgent-Next: A Multimodal Generalist Agent for Automated Computer Interaction}

\author{
    \textbf{Bin Lei}$^{1}$\thanks{Equal Contribution.}{ } \textbf{ Weitai Kang}$^{2*}${ } \textbf{Zijian Zhang}$^{1}${ } \textbf{Winson Chen}$^1${ } \textbf{Xi Xie}$^{3}${ } \textbf{Shan Zuo}$^{3}${ } \\ \textbf{Mimi Xie}$^{4}${ } \textbf{Ali Payani}$^{5}${ } \textbf{Mingyi Hong}$^{1}${ } \textbf{Yan Yan}$^{2}${ } \textbf{Caiwen Ding}$^{1}${ }\\
    $^1$University of Minnesota { }
    $^2$University of Illinois Chicago { } \\
    $^3$University of Connecticut { }
    $^4$The University of Texas at San Antonio
    $^5$Cisco Research \\
    \texttt{lei00126@umn.edu}, \texttt{wkang126@uic.edu}
}

\begin{document}
\maketitle
\input{Sections/Abstract}

\input{Sections/Introduction}
\input{Sections/Related_Work}
\input{Sections/Infantagent_Next}
\input{Sections/Experiment}
\input{Sections/Conclusion}
\input{Sections/Acknowledgements}
{\small
\bibliographystyle{plain}
\bibliography{reference}
}

\input{Sections/Appendix}
\end{document}

%% file: Sections/Abstract.tex
\begin{abstract}
This paper introduces \textsc{InfantAgent-Next}, a generalist agent capable of interacting with computers in a multimodal manner, encompassing text, images, audio, and video.
Unlike existing approaches that either build intricate workflows around a single large model or only provide workflow modularity, our agent integrates tool-based and pure vision agents within a highly modular architecture, enabling different models to collaboratively solve decoupled tasks in a step-by-step manner. 
Our generality is demonstrated by our ability to evaluate not only pure vision-based real-world benchmarks (i.e., OSWorld), but also more general or tool-intensive benchmarks (e.g., GAIA and SWE-Bench).
Specifically,
we
achieve a $\mathbf{7.27\%}$ accuracy gain over Claude-Computer-Use on OSWorld. 
Codes and evaluation scripts are 
open-sourced at \url{https://github.com/bin123apple/InfantAgent}.
\end{abstract}

%% file: Sections/Introduction.tex
\section{Introduction}

Automated AI agents~\cite{lei2024infant, sager2025ai, ning2025survey, wang2024openhands, li2023camel, hong2023metagpt} are vital in today’s digital era. By integrating large language models (LLMs)~\cite{achiam2023gpt, guo2025deepseek} and visual large language models (vLLMs)~\cite{chatgpt, claude, llava1.5, qwen2, qwen2.5}, they can multimodally understand user intents—including text, images, voice, and video—and, with minimal human intervention, convert these intents into precise sequences of interface actions. Through LLM / vLLM-driven planning, advanced perception of UI elements (such as buttons, text fields, and images), and a modular execution framework, they significantly reduce manual effort and enable the efficient execution of complex workflows across diverse software environments.

Current automated AI agents can be broadly classified into two categories, and they have not made sufficiently fine-grained distinctions at the agent's tool selection and execution levels, which leads to significant limitations in practice.
\raisebox{-1.1pt}{\ding[1.1]{182\relax}}
The first category consists of \textit{tool-based agents}, such as OpenHands~\cite{wang2024openhands}, OWL~\cite{li2023camel}, and AutoGPT~\cite{Significant-Gravitas_AutoGPT_2025}, that equip LLMs with a suite of predefined tools (e.g., code generation tools, web search tools) to boost task-specific accuracy. However, because these agents usually rely on a single model to decide when and how to use each tool, they require the manual definition and integration of tools for every possible desktop scenario—an infeasible and brittle approach that limits generality. 
\raisebox{-1.1pt}{\ding[1.1]{183\relax}}
The second category comprises \textit{pure vision-based agents}, such as UI-TARS~\cite{qin2025ui}, and Aguvis~\cite{xu2024aguvis}, which use vLLMs to control computers via GUI. This design allows broader applicability, since it bypasses the need for tool integration. However, high-resolution reasoning with a single model hurts accuracy on tasks that could be easily handled by simple tool calls, such as document editing or code manipulation—where tool-based agents excel.
\raisebox{-1.1pt}{\ding[1.1]{184\relax}}
Although advanced models (e.g., Claude-3.7-Sonnet~\cite{claude}, o3~\cite{o3}) can decompose complex problems into precise, step‐by‐step plans, they frequently fail at execution—mislocalizing GUI click coordinates~\cite{cheng2024seeclick} (GPT-4o~\cite{hurst2024gpt} attains only $\mathbf{0.8\%}$ accuracy on the ScreenSpot-Pro benchmark~\cite{li2025screenspot}) or selecting incorrect line numbers during file edits. Conversely, specialized vision modules (such as visual-grounding models~\cite{lin2024showui, gou2024navigating}) exhibit limited reasoning capacity and restricted context windows, preventing reliable inference of subsequent actions.
Ensuring both high task‐level accuracy and broad generality requires a hybrid agent paradigm that unifies tool‐based and pure‐vision approaches. 

\begin{figure}[ht]
\centering
\includegraphics[width=0.95\textwidth]{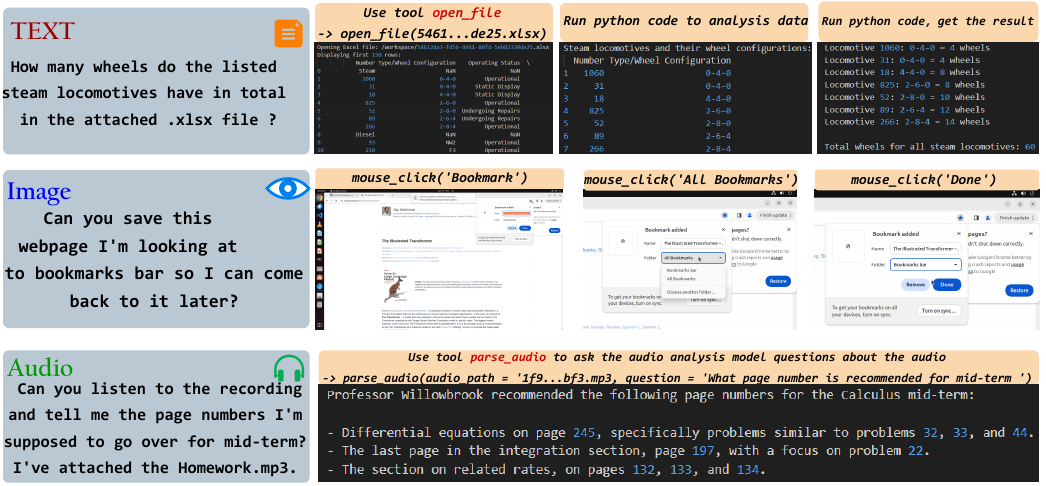}
\caption{Three real-world task examples addressed by \textsc{InfantAgent-Next}, each requiring different modality capabilities.
Input requests are shown in the \includegraphics[height=0.7em]{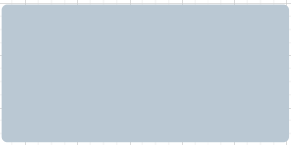} block, actions taken by the agent are shown in the \includegraphics[height=0.7em]{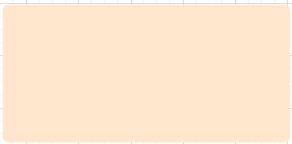} block, and execution results appear directly below the action block.}
\label{fig:Use_cases}
\end{figure}

Inspired by these challenges, we present \textsc{InfantAgent-Next}, which undertakes detailed modularization of agent workflows, tool selection, and tool execution, in favor of a modular architecture with a unified dialogue context. Each subtask is routed to the most appropriate specialist—reasoning models handle logical inference, visual-grounding models localize UI elements, audio-analysis models interpret sound, and so on—and their outputs are seamlessly merged back into the conversation history. This design enables true multimodal interaction with computer interfaces, rather than being confined to formatted HTML or Accessibility Tree manipulations or purely vision-based control. Figure~\ref{fig:Use_cases} presents several real-world task examples addressed by \textsc{InfantAgent-Next}.

Our contributions are as follows: 
\textbf{(i)} We introduce \textsc{InfantAgent-Next}, an open-source, multimodal generalist agent framework that interacts with its host computer via text, images, and audio.
\textbf{(ii)}
We have modularized the agent’s workflow, tool selection, and tool execution in detail—unifying the tool-based and pure-vision agent paradigms to achieve both high task-level accuracy and broad generality.
\textbf{(iii)}
We optimize a suite of commonly used agent tools and release them as open-source which supports isolated \textcolor{red}{} execution environment, enabling future research in the community.

%% file: Sections/Related_Work.tex
\section{Related Work Comparison}
\vspace{-4pt}
In Table~\ref{tab:Related_Work}, we compare \textsc{InfantAgent-Next} with existing agents across eight key dimensions. 

\noindent{\textbf{\textit{Generalist:}}} Generalist agents are capable of supporting various computer tasks, while specialist agents are typically designed to address a specific class of problems. For example, OpenHands~\cite{wang2024openhands} and MetaGPT~\cite{hong2023metagpt} primarily focus on software engineering tasks.


\noindent{\textbf{\textit{Built-in Tools:}}} For common operations such as web search or file editing, integrating built-in tools into agents can greatly streamline workflows and enable structured output generation. For instance, AutoGPT~\cite{Significant-Gravitas_AutoGPT_2025}’s \texttt{search\_web} tool retrieves search results by issuing HTTP requests and parsing the returned JSON/HTML structure. In contrast, agents like UI-TARS~\cite{qin2025ui} lack such tools and rely solely on pure vision-based capabilities for parsing.

\noindent{\textbf{\textit{Visual Grounding:}}} 
Built-in tools are often insufficient for diverse computer softwares—particularly with professional graphics software or niche PC games. Agents such as ShowUI~\cite{lin2024showui}, which are equipped with visual grounding capabilities, are better suited for handling these scenarios.

\noindent{\textbf{\textit{Multi-Model Support:}}} 
Different models exhibit different strengths: reasoning models are effective at complex analytical tasks, non-reasoning models offer faster response times, audio analysis models handle auditory inputs, and visual grounding models enable precise spatial localization. Assigning each model to tasks that match its strengths can significantly enhance overall agent performance. MetaGPT~\cite{hong2023metagpt} exemplifies this multi-model integration strategy.

\noindent{\textbf{\textit{Memory Retrieval:}}} 
As dialogue histories grow, model inference overhead increases. This overhead can be mitigated by retrieving relevant memory segments and injecting them into the context on demand. LangChain~\cite{langchain} demonstrates this approach by offering multiple memory classes—such as \texttt{summary memory} and \texttt{retrieval-based memory}.

\noindent{\textbf{\textit{Dynamic Toolset:}}} 
To extend an agent’s functionality, developers often include a broad range of tools. Tool usage instructions are typically prepended to the conversation history to guide the LLM. However, a large tool inventory can lead to increased inference cost and challenging tool selection. Frameworks like LangChain~\cite{langchain} address this by dynamically selecting a relevant subset of tools from the full library, thereby reducing the cognitive and computational burden.

\noindent{\textbf{\textit{User Interaction during Execution:}}} 
User’s initial prompts are sometimes underspecified (e.g., \textit{Please write a short story.}). In such cases, it is crucial for the agent to interactively elicit additional details rather than completing the task blindly. For example, OpenHands~\cite{wang2024openhands} includes a \texttt{MessageAction} class that enables the agent to request clarification from the user.

\noindent{\textbf{\textit{Dedicated Computer:}}} 
We design a fully isolated execution environment supporting CLI, Python scripting, and GUI interaction; implementation details are provided in the Appendix~\ref{Dedicated}.

\setlength{\intextsep}{0pt}

\begin{table}[h]
\centering
\caption{Comparison of \textsc{InfantAgent-Next} with Different Agent Frameworks}
\label{tab:Related_Work}
\begin{adjustbox}{width=0.99\textwidth}
\begin{tabular}{c|cccccccc}
\hline
Name             & \begin{tabular}[c]{@{}c@{}}Generalist \\ Agent\end{tabular} & \begin{tabular}[c]{@{}c@{}}Built-in \\ Tools\end{tabular} & \begin{tabular}[c]{@{}c@{}}Visual \\ Grounding\end{tabular} & \begin{tabular}[c]{@{}c@{}}Multi-Model \\ Support\end{tabular} & \begin{tabular}[c]{@{}c@{}}Memory \\ Retrieval\end{tabular} & \begin{tabular}[c]{@{}c@{}}Dynamic \\ Toolset\end{tabular} & \begin{tabular}[c]{@{}c@{}}User Interaction \\ during Execution\end{tabular} & \begin{tabular}[c]{@{}c@{}}Dedicated  \\ Computer\end{tabular} \\ \hline
AutoGPT~\cite{Significant-Gravitas_AutoGPT_2025}          & \textcolor{darkgreen}{\ding{51}}                                                                & \textcolor{darkgreen}{\ding{51}}                                                                  &         \textcolor{red}{\ding{55}}                                                    &  \textcolor{darkgreen}{\ding{51}}                                                               &    \textcolor{red}{\ding{55}}                                                              &      \textcolor{red}{\ding{55}}                                                      &      \textcolor{red}{\ding{55}}                                                                        &         \textcolor{red}{\ding{55}}                                                                   \\
BabyAGI~\cite{babyagi}          & \textcolor{darkgreen}{\ding{51}}                                                                  & \textcolor{red}{\ding{55}}                                                                &    \textcolor{red}{\ding{55}}                                                         &    \textcolor{red}{\ding{55}}                                                            &          \textcolor{red}{\ding{55}}                                                        &     \textcolor{red}{\ding{55}}                                                       &        \textcolor{red}{\ding{55}}                                                                      &                                                                     \textcolor{red}{\ding{55}}       \\
LangChain~\cite{langchain}        & \textcolor{darkgreen}{\ding{51}}                                                                 & \textcolor{darkgreen}{\ding{51}}                                                                &     \textcolor{red}{\ding{55}}                                                        &   \textcolor{darkgreen}{\ding{51}}                                                             &              \textcolor{darkgreen}{\ding{51}}                                                    &   \textcolor{darkgreen}{\ding{51}}                                                        &       \textcolor{darkgreen}{\ding{51}}                                                                       &   \textcolor{red}{\ding{55}}                                                                         \\
MetaGPT~\cite{hong2023metagpt}          & \textcolor{red}{\ding{55}}                                                                & \textcolor{darkgreen}{\ding{51}}                                                                &  \textcolor{red}{\ding{55}}                                                           &                                             \textcolor{darkgreen}{\ding{51}}                   &       \textcolor{red}{\ding{55}}                                                           &    \textcolor{red}{\ding{55}}                                                        &         \textcolor{red}{\ding{55}}                                                                     &                                                                    \textcolor{red}{\ding{55}}        \\
OpenHands~\cite{wang2024openhands}        & \textcolor{red}{\ding{55}}                                                               & \textcolor{darkgreen}{\ding{51}}                                                               &  \textcolor{red}{\ding{55}}                                                           &                            \textcolor{darkgreen}{\ding{51}}                                    &   \textcolor{red}{\ding{55}}                                                               &   \textcolor{red}{\ding{55}}                                                         &          \textcolor{darkgreen}{\ding{51}}                                                                    &                    \textcolor{red}{\ding{55}}                                                        \\
UI-TARS~\cite{qin2025ui}          & \textcolor{red}{\ding{55}}                                                                  & \textcolor{red}{\ding{55}}                                                                &       \textcolor{darkgreen}{\ding{51}}                                                      &      \textcolor{red}{\ding{55}}                                                           &   \textcolor{red}{\ding{55}}                                                                &   \textcolor{red}{\ding{55}}                                                          &    \textcolor{red}{\ding{55}}                                                                           &     \textcolor{red}{\ding{55}}                                                                        \\
AutoGen~\cite{wu2023autogen}          & \textcolor{darkgreen}{\ding{51}}                                                                  & \textcolor{darkgreen}{\ding{51}}                                                                &  \textcolor{red}{\ding{55}}                                                            &                              \textcolor{darkgreen}{\ding{51}}                                  &  \textcolor{red}{\ding{55}}                                                                 &  \textcolor{red}{\ding{55}}                                                           &           \textcolor{red}{\ding{55}}                                                                    &                      \textcolor{red}{\ding{55}}                                                       \\
AGUVIS~\cite{xu2024aguvis}           & \textcolor{red}{\ding{55}}                                                                 & \textcolor{red}{\ding{55}}                                                                &      \textcolor{darkgreen}{\ding{51}}                                                       &      \textcolor{red}{\ding{55}}                                                          &   \textcolor{red}{\ding{55}}                                                               &    \textcolor{red}{\ding{55}}                                                        &   \textcolor{red}{\ding{55}}                                                                           &    \textcolor{red}{\ding{55}}                                                                        \\
ShowUI~\cite{lin2024showui}           & \textcolor{red}{\ding{55}}                                                                  & \textcolor{red}{\ding{55}}                                                                &        \textcolor{darkgreen}{\ding{51}}                                                     &    \textcolor{red}{\ding{55}}                                                             &  \textcolor{red}{\ding{55}}                                                                  &     \textcolor{red}{\ding{55}}                                                         &    \textcolor{red}{\ding{55}}                                                                            &   \textcolor{red}{\ding{55}}                                                                           \\
OS-Atlas~\cite{wu2024atlas}         & \textcolor{red}{\ding{55}}                                                                & \textcolor{red}{\ding{55}}                                                                &       \textcolor{darkgreen}{\ding{51}}                                                      &    \textcolor{darkgreen}{\ding{51}}                                                            &     \textcolor{red}{\ding{55}}                                                             &    \textcolor{red}{\ding{55}}                                                        &                     \textcolor{red}{\ding{55}}                                                         &     \textcolor{red}{\ding{55}}                                                                       \\
AutoAgents~\cite{chen2023autoagents}       & \textcolor{darkgreen}{\ding{51}}                                                                 & \textcolor{red}{\ding{55}}                                                                 &            \textcolor{red}{\ding{55}}                                                  & \textcolor{darkgreen}{\ding{51}}                                                               &   \textcolor{red}{\ding{55}}                                                                &   \textcolor{darkgreen}{\ding{51}}                                                         &        \textcolor{red}{\ding{55}}                                                                       &              \textcolor{red}{\ding{55}}                                                               \\ 
\hline
\textbf{InfantAgent-Next} & \textcolor{darkgreen}{\ding{51}}                                                                  &  \textcolor{darkgreen}{\ding{51}}                                                                 &   \textcolor{darkgreen}{\ding{51}}                                                           &              \textcolor{darkgreen}{\ding{51}}                                                   &      \textcolor{darkgreen}{\ding{51}}                                                             &    \textcolor{darkgreen}{\ding{51}}                                                         &                            \textcolor{darkgreen}{\ding{51}}                                                   &                       \textcolor{darkgreen}{\ding{51}}                                                      \\ \hline
\end{tabular}
\end{adjustbox}
\vspace{4pt}
\end{table}
\setlength{\intextsep}{0pt}
Compared to prior agent frameworks, \textsc{InfantAgent-Next} uniquely integrates all of the above features. See Section~\ref{sec:Infantagent_Next} for further details.

%% file: Sections/Infantagent_Next.tex
\section{InfantAgent-Next}\label{sec:Infantagent_Next}
In this section, we first introduce the overall architecture of \textsc{InfantAgent-Next} to illustrate its operational logic in Sec.~\ref{architecture}. We then describe some crucial components in Sec.~\ref{component}, including the memory extraction mechanism employed during execution, toolkit configured for the system, etc. Finally, we propose our novel improvement on important actions of the agent in Sec.~\ref{mouse_file}, including Mouse Click and File Edit. We provide a detailed case study in Appendix~\ref{apd:case_analysis}.

\subsection{Architecture}\label{architecture}
\begin{figure}[]
  \centering
  \includegraphics[width=0.95\textwidth]{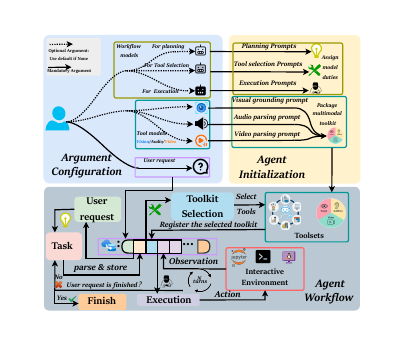} 
\caption{\textsc{InfantAgent-Next} architecture overview.
\includegraphics[height=0.7em]{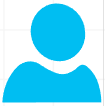}: User input argument and request. \textbf{Environment related icons:} \includegraphics[height=0.7em]{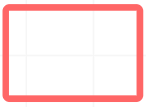}: Agent Interaction Environment. \includegraphics[height=0.7em]{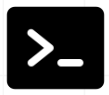}: Terminal interface. \includegraphics[height=0.7em]{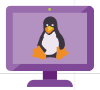}: GNOME desktop. \includegraphics[height=0.7em]{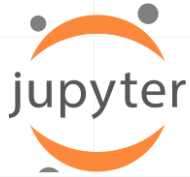}: Jupyter. \textbf{Models related icons:} 
\includegraphics[height=0.7em]{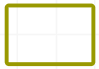}: Load Workflow models.
\includegraphics[height=0.7em]{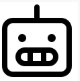}: Planning model
\includegraphics[height=0.7em]{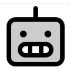}: Tool Selection model
\includegraphics[height=0.7em]{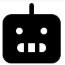}: Execution model \includegraphics[height=0.7em]{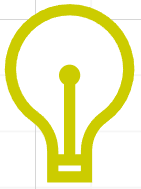}: Planner. \includegraphics[height=0.7em]{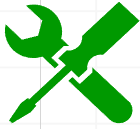}: Tool Selector. \includegraphics[height=0.7em]{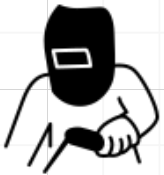}: Executer. 
\textbf{Tools related icon:} \includegraphics[height=0.7em]{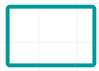}: Load Toolsets. \includegraphics[height=0.7em]{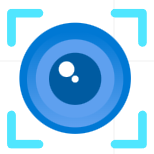}: Tool models argument: \texttt{Vision\_model\_name}. \includegraphics[height=0.7em]{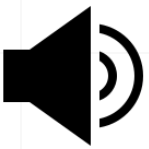}: Tool models argument: \texttt{Audio\_model\_name}. \includegraphics[height=0.7em]{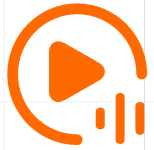}: Tool models argument: \texttt{Video\_model\_name}. \includegraphics[height=0.7em]{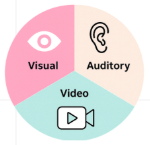}: Multimodal toolkit. \includegraphics[height=0.7em]{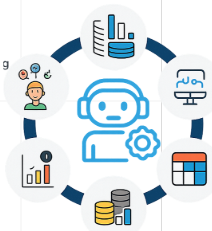}: Other Toolkits.
\textbf{Memory related icons:} \includegraphics[height=0.7em]{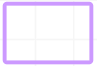}: Load agent memory. \includegraphics[height=0.7em]{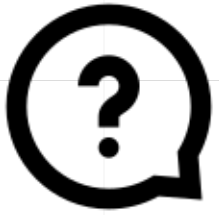}: Request from the user. \includegraphics[height=0.7em]{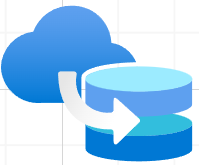}: Agent Memory Cache, used for storing all memories.}
  \label{fig:architecture_overview}
  \vspace{-10pt}
\end{figure}

\noindent\textbf{\textit{Overview:}} The architecture of \textsc{InfantAgent-Next} is illustrated in Figure~\ref{fig:architecture_overview}. \textsc{InfantAgent-Next} meticulously modularizes the workflow where different models complete different steps.
The process begins when the user submits a request. We first allow the user to configure the arguments for Workflow models and Tool models. 
After the agent is initialized, 
it analyzes the user’s request to generate tasks. It then performs tool selection and invokes the selected tool to complete the tasks. Finally, the agent checks whether the current task has been successfully completed.

\textbf{\uppercase\expandafter{\romannumeral 1}}. \textit{Argument Configuration} (Figure~\ref{fig:architecture_overview}-upper left): In this stage, users can customize and assign different models to different tasks. These include three workflow models responsible for planning, tool selection, and task execution, respectively, and three modality-specific tool models for handling images, audio, and video. 

\textbf{\uppercase\expandafter{\romannumeral 2}}. \textit{Agent Initialization} (Figure~\ref{fig:architecture_overview}-upper right):
During initialization, we assign specific roles to each model with tailored prompt templates. 
For example, some tools are associated with specific tool models and their prompt templates, and are then packaged into a unified multimodal toolkit.
At this stage, the agent’s memory cache is initialized, with the user’s request stored as the first entry.

\textbf{\uppercase\expandafter{\romannumeral 3}}. \textit{Agent Workflow} (Figure~\ref{fig:architecture_overview}-bottom): The initialized multimodal toolkits, together with a suite of predefined utility functions, are organized into a comprehensive toolset. This set of tools, along with the interactive environment, all models, and the agent's memory cache, is integrated into the agent’s workflow. 
The user’s initial request serves as the input and triggers the following iterative loop:
\raisebox{-1.1pt}{\ding[1.1]{182\relax}} 
The Planner model analyzes the request and current state to produce a task, which is parsed and stored in memory.
\raisebox{-1.1pt}{\ding[1.1]{183\relax}} 
The agent reconstructs the memory using a prompt template for toolkit selection. The Tool Selection model then selects the most appropriate tool from the toolset. The selected tool is registered and activated for this iteration, while others remain idle.
\raisebox{-1.1pt}{\ding[1.1]{184\relax}} 
The Executor model invokes the tools and gathers feedback from the environment to complete the task.
\raisebox{-1.1pt}{\ding[1.1]{185\relax}} 
If we successfully resolve the task, the loop terminates; otherwise, the process proceeds to the next iteration.




\subsection{Component Design}\label{component}
\noindent{\textbf{\textit{Memory:}}} 
Since the agent must perform three distinct types of tasks—planning, tool selection, and execution—it is insufficient to simply log the raw dialogue history. 
Instead, each model-generated response is parsed using special tags and sequentially stored in the agent’s memory cache.
The agent’s reasoning is enclosed within \texttt{<thought>...</thought>} tags; task assignments are marked by \texttt{<task>...</task>}; selected toolkits are indicated by \texttt{<toolkit>...</toolkit>}; and tool executions are represented by either \texttt{<execute\_bash>...</execute\_bash>} or \texttt{<execute\_python>...</execute\_python>}.
When the agent transitions to a new task, a subset of the memory cache is extracted and reconstructed into the dialogue by concatenating relevant memory attributes. 
Task-specific tags are activated to prevent the LLM from generating irrelevant content. Finally, the reconstructed dialogue is combined with predefined prompt templates to form the conversation history for that task. The restoration process follows these steps:

\textbf{\uppercase\expandafter{\romannumeral 1}}. \textit{Planning}: During the planning phase, the agent retrieves all memories except those associated with tool selection. All tags are removed during reconstruction, except for the \texttt{<task>...</task>} tags. These reconstructed memories are inserted between the planning system prompt and the end prompt for planning, both of which are detailed in Appendix~\ref{apd:planning_prompt_template}.

\textbf{\uppercase\expandafter{\romannumeral 2}}. \textit{Tool selection}: 
For tool selection, only the most recent task memory is retrieved from the cache, and all associated tags are removed. This content is then appended directly after the tool selection system prompt, as provided in Appendix~\ref{apd:tool_selection_prompt_template}.

\textbf{\uppercase\expandafter{\romannumeral 3}}. \textit{Execution}: 
During execution, all memories except those related to tool selection are retained. When reconstructing them into dialogue, only the \texttt{<execute\_bash>...</execute\_bash>} and \texttt{<execute\_python>...</execute\_python>} tags are preserved. The resulting dialogue is inserted between the execution system prompt and execution end prompt, shown in Appendix~\ref{apd:execution_prompt_template}.

\noindent{\textbf{\textit{Tools:}}} 
To invoke a tool, the agent prepends its documentation and usage examples to the dialogue history. 
However, as the toolset expands, this strategy introduces two challenges: (1) inflated inference overhead due to the volume of text, and (2) increased difficulty for the model to identify the correct tool. 
To mitigate these issues, we categorize tools and dynamically select a relevant subset during each tool selection step. Specifically, we define seven toolkits: File Reading, File Searching, File Editing, Web Browsing, Computer Use, Code Execution, and Advanced Tools. The Computer Use toolkit includes all keyboard and mouse operations as well as multimodal tools, while the Advanced Tools toolkit provides various composite commands. Each toolkit is accompanied by a usage example. Detailed descriptions of tool functions are available in Appendix~\ref{apd:Toolkits}.

\subsection{Mouse Click \& File Edit}\label{mouse_file}
As noted earlier, SOTA models, such as Claude-3.7-Sonnet or GPT-4o, remain limited in mouse-click action (visual grounding) accuracy and robust file editing. We incorporate dedicated mechanisms to mitigate these limitations.

\begin{wrapfigure}{r}{0.77\textwidth}   
  \begin{minipage}{0.77\textwidth}
    \begin{algorithm}[H]
      \caption{Iterative Region Cropping and Mouse Click Logic}
      \label{alg:iterative_mouse_click}
      \begin{algorithmic}[1]
        \Require \textit{agent}
        \If{$\text{Last}(\textit{agent.memory}).\text{type} \neq \texttt{"mouse\_click"}$}
          \State \Return
        \EndIf
        \State $description \gets \text{Last}(\textit{agent.memory}).description$
        \State $region \gets screenshot$
        \For{$i = 1$ \textbf{to} $n$}
          \State $coord_i \gets \Call{VisionGrounding}{description, region}$
          \State $region \gets \Call{CropScreen}{center=coord_i,\; size=\text{CROP\_SIZE}}$
        \EndFor
        \State $final\_coord \gets \Call{VisionGrounding}{description,\; region}$
        \State \Call{PerformClick}{final\_coord}
      \end{algorithmic}
    \end{algorithm}
  \end{minipage}
  \vspace{-10pt}
\end{wrapfigure}

\setlength{\intextsep}{0pt}

\noindent{\textbf{\textit{Mouse Click:}}}
For \texttt{mouse\_click} operations, the model is required to output two parameters: (1) the name of the target element (e.g., Google Chrome or VS Code icon), and (2) a detailed description of its position and shape. 
As illustrated in Algorithm~\ref{alg:iterative_mouse_click}, we first verify whether the agent’s most recent action is \texttt{mouse\_click}; if not, the procedure terminates early. 
Otherwise, we retrieve the associated description from memory and treat a full-screen screenshot as the initial search region. 
We then enter an iterative loop of length $n$, where each iteration passes the current region and the description to the visual grounding model to obtain a candidate coordinate $coord_i$. 
We subsequently call \texttt{CropScreen} centered at $coord_i$ to extract a smaller region for the next step. 
After completing $n$ iterations, the final region is passed to the visual grounding model once more to predict the refined click position $final\_coord$, which is then used by \texttt{PerformClick} to click on the screen. 
By progressively narrowing the search region, this method enhances visual grounding precision and reduces distraction from irrelevant pixels.

\begin{wrapfigure}{r}{0.63\textwidth}   
  \begin{minipage}{0.63\textwidth}      
    \begin{algorithm}[H]
      \caption{File Editing Logic}
      \label{alg:simple_file_edit}
      \begin{algorithmic}[1]
        \Require \textit{agent}
        \If{$\text{Last}(\textit{agent.memory}).\text{type} \neq \texttt{"file\_edit"}$}
          \State \Return
        \EndIf
        \State $request \gets \text{Last}(\textit{agent.memory}).\text{edit\_details}$
        \For{$i \gets 1$ \textbf{to} MAX\_ITER}
          \LineComment{Generate edit plan}
          \State $(s, s_{\text{cont}}, e, e_{\text{cont}}) \gets \Call{GenerateEditPlan}{request}$
          \LineComment{Verify plan boundaries}
          \If{$\text{FileLine}(s)=s_{\text{cont}} \,\wedge\, \text{FileLine}(e)=e_{\text{cont}}$}
            \State \Call{ApplyEdit}{s, e, request}
            \State \Return
          \EndIf
          \LineComment{Fallback: locate best matching lines}
          \State $a_s \gets \text{FileLine}(s)$ 
          \State $a_e \gets \text{FileLine}(e)$
          \State $m_s \gets \Call{Fuzzify}{s_{\text{cont}}}$
          \State $m_e \gets \Call{Fuzzify}{e_{\text{cont}}}$
          \State $(s', e') \gets \Call{FindBestMatch}{m_s, m_e}$
          \State $request \gets (s', e', a_s, a_e, request)$
        \EndFor
      \end{algorithmic}
    \end{algorithm}
  \end{minipage}
\end{wrapfigure}
\noindent{\textbf{\textit{File Editing:}}}
There are two main formats for the \texttt{file\_edit} command.
(1) The three-parameter format used by SWE-Agent~\cite{SWE-agent} specifies the file path, the exact string to replace, and the replacement. It often fails when the target string is long or occurs multiple times.
(2) The four-parameter format used by OpenHands specifies the file path, start and end line numbers, and the replacement string, but is prone to errors such as incorrect or overlapping line ranges.
To ensure that the agent can perform file edits accurately, we have encapsulated the logic from Algorithm~\ref{alg:simple_file_edit} as follows:
\raisebox{-1.1pt}{\ding[1.1]{182\relax}} First, we verify whether the agent’s most recent memory entry is of type \texttt{file\_edit}. If not, the procedure terminates immediately.
\raisebox{-1.1pt}{\ding[1.1]{183\relax}} If it is, the agent extracts the edit request from memory and invokes \texttt{GenerateEditPlan(request)}, which returns the proposed starting and ending line numbers $s$ and $e$, along with the expected contents of those lines, $s\_cont$ and $e\_cont$.
\raisebox{-1.1pt}{\ding[1.1]{184\relax}} The agent then checks whether the actual content of line $s$ matches $s\_cont$, and whether line $e$ matches $e\_cont$. If both checks pass, it directly applies the edit using \texttt{ApplyEdit(s, e, request)} and terminates.
\raisebox{-1.1pt}{\ding[1.1]{185\relax}} If the boundary checks fail, the agent switches to a fallback strategy: it first saves the original contents at lines $s$ and $e$ as $a\_s$ and $a\_e$, then fuzzifies the expected snippets via \texttt{m\_s = Fuzzify(s\_cont), m\_e = Fuzzify(e\_cont)} and calls \texttt{FindBestMatch(m\_s, m\_e)} to locate the most similar span ($s'$, $e'$) in the file. The edit request is then updated with the new span and original content $(s', e', a\_s, a\_e, request)$, and the process repeats for up to \texttt{MAX\_ITER} iterations.

%% file: Sections/Experiment.tex
\vspace{-10pt}
\section{Experiment}
\subsection{Vision Ability}
\noindent{\textbf{\textit{Benchmark.}}} 
We assess the visual reasoning capability of \textsc{InfantAgent-Next} using the OSWorld benchmark~\cite{OSWorld}.
OSWorld provides a scalable, real-world computer environment comprising 369 open-ended desktop tasks. 
Each task is initialized with a fully specified machine state, including a high-resolution screenshot, active application windows, and file system context, paired with a natural language instruction and an executable evaluation script. 
The tasks span diverse domains, such as web browsing, file operations, code editing, image manipulation, and multi-application workflows, challenging the agent's ability to ground natural language instructions in GUI elements.

\noindent{\textbf{\textit{Experiment Setup.}}} 
The evaluation is conducted using the official OSWorld codebase~\cite{OSWorld}, executed within a VMware virtual machine. 
Mouse click events are simulated using \texttt{PyAutoGUI}. For reasoning, we use \texttt{Claude-3.7-Sonnet}, while visual grounding is performed using \texttt{UI-TARS-1.5-7B}, with \texttt{max\_steps} set to $50$. For fair comparison with baseline methods, all non-visual toolkits are disabled.

\noindent{\textbf{\textit{Results.}}} 
As presented in Table~\ref{tab:OSWorld}, \textsc{InfantAgent-Next} achieves superior accuracy at $50$ steps compared to OpenAI CUA and Claude Computer Use. When utilizing \texttt{Claude-3.7-Sonnet} as the planner model, \textsc{InfantAgent-Next} outperforms Agent-S2~\cite{agashe2025agent} and all other open-source frameworks, demonstrating its enhanced visual reasoning capabilities.

\begin{table}[t]
\centering
\caption{Performance of \textsc{InfantAgent-Next} on OSWorld. {\small \textcolor{red}{\faLock}}: Close source. \cmark: Open source.}
\label{tab:OSWorld}
\begin{adjustbox}{max width=\textwidth}
\begin{tabular}{ccccc}
\hline
\rowcolor[HTML]{FFFFFF} 
{\color[HTML]{222222} Framework}           & {\color[HTML]{222222} Model}                              & {\color[HTML]{222222} Steps} & {\color[HTML]{222222} Open Source} & {\color[HTML]{222222} Accuracy} \\ \hline
\rowcolor[HTML]{FFFFFF} 
{\color[HTML]{222222} UI-TARS~\cite{qin2025ui}}             & {\color[HTML]{222222} UI-TARS-1.5-72B}                    & {\color[HTML]{222222} $100$}   & {\color[HTML]{222222} {\small \textcolor{red}{\faLock}}}           & {\color[HTML]{222222} $42.5$}     \\
\rowcolor[HTML]{FFFFFF} 
{\color[HTML]{222222} OpenAI CUA~\cite{CUA}}          & {\color[HTML]{222222} GPT-4o}                             & {\color[HTML]{222222} $200$}   & {\color[HTML]{222222} {\small \textcolor{red}{\faLock}}}           & {\color[HTML]{222222} $38.1$}     \\
\rowcolor[HTML]{FFFFFF} 
{\color[HTML]{222222} UI-TARS~\cite{qin2025ui}}             & {\color[HTML]{222222} UI-TARS-1.5-72B}                    & {\color[HTML]{222222} $50$}    & {\color[HTML]{222222} {\small \textcolor{red}{\faLock}}}           & {\color[HTML]{222222} $38.0$}     \\
\rowcolor[HTML]{C0C0C0} 
{\color[HTML]{222222} InfantAgent-Next}    & {\color[HTML]{222222} Claude-3.7-Sonnet + UI-TARS-1.5-7B} & {\color[HTML]{222222} $50$}    & {\color[HTML]{222222} \cmark}           & {\color[HTML]{222222} $35.3$}    \\
\rowcolor[HTML]{FFFFFF} 
{\color[HTML]{222222} Agent S2~\cite{agashe2025agent}}            & {\color[HTML]{222222} Claude-3.7-Sonnet}                  & {\color[HTML]{222222} $50$}    & {\color[HTML]{222222} \cmark}           & {\color[HTML]{222222} $34.5$}     \\
\rowcolor[HTML]{FFFFFF} 
{\color[HTML]{222222} OpenAI CUA~\cite{CUA}}          & {\color[HTML]{222222} GPT-4o}                             & {\color[HTML]{222222} $50$}    & {\color[HTML]{222222} {\small \textcolor{red}{\faLock}}}           & {\color[HTML]{222222} $32.6$}     \\
\rowcolor[HTML]{FFFFFF} 
{\color[HTML]{222222} Claude Computer Use~\cite{modelsandcomputeruse}} & {\color[HTML]{222222} Claude-3.7-Sonnet}                  & {\color[HTML]{222222} $100$}   & {\color[HTML]{222222} {\small \textcolor{red}{\faLock}}}           & {\color[HTML]{222222} $28.0$}     \\
\rowcolor[HTML]{FFFFFF} 
{\color[HTML]{222222} UI-TARS~\cite{qin2025ui}}             & {\color[HTML]{222222} UI-TARS-1.5 7B}                     & {\color[HTML]{222222} $100$}   & {\color[HTML]{222222} \cmark}           & {\color[HTML]{222222} $26.9$}    \\
\rowcolor[HTML]{FFFFFF} 
{\color[HTML]{222222} Claude Computer Use~\cite{modelsandcomputeruse}} & {\color[HTML]{222222} Claude-3.7-Sonnet}                  & {\color[HTML]{222222} $50$}    & {\color[HTML]{222222} {\small \textcolor{red}{\faLock}}}           & {\color[HTML]{222222} $26.0$}       \\
\rowcolor[HTML]{FFFFFF} 
{\color[HTML]{222222} UI-TARS~\cite{qin2025ui}}             & {\color[HTML]{222222} UI-TARS-72B-DPO}                    & {\color[HTML]{222222} $50$}    & {\color[HTML]{222222} \cmark}           & {\color[HTML]{222222} $24.6$}     \\
\rowcolor[HTML]{FFFFFF} 
{\color[HTML]{222222} AGUVIS~\cite{xu2024aguvis}}              & {\color[HTML]{222222} AGUVIS 72B + GPT-4o}                & {\color[HTML]{222222} -}     & {\color[HTML]{222222} \cmark}           & {\color[HTML]{222222} $17.0$}    \\
Aria-UI~\cite{yang2024aria}                                    & Aria-UI + GPT-4o                                          & -                            & \cmark                                  & $15.1$                            \\
OS-Atlas~\cite{wu2024atlas}                                   & OS-Atlas-Base-7B + GPT-4o                                 & -                            & \cmark                                  & $14.6$                           \\
SeeClick~\cite{cheng2024seeclick}                                   & SeeClick + GPT-4o                                         & -                            & \cmark                                  & 9.21                            \\
Qwen2.5~\cite{qwen2.5}                                    & Qwen2.5-vl-72B                                            & -                            & \cmark                                  & $8.8$                             \\ \hline
\end{tabular}
\end{adjustbox}
\end{table}


\subsection{Logic Reasoning Ability}

\begin{table}[t]
\centering
\caption{Performance on SWE-Bench-Verified and SWE-Bench-Lite.}
\label{tab:GAIA}
\begin{adjustbox}{max width=\textwidth}
\begin{tabular}{ccc}
\hline
Open/Close Source & Method & Accuracy (\%) \\
\hline
\multicolumn{3}{c}{\textbf{SWE-Bench-Verified (50 cases)}} \\
\hline
\multicolumn{1}{c|}{} & SWE-agent + Claude 3.7 Sonnet w/ Review Heavy~\cite{SWE-agent} & $72$ \\
\multicolumn{1}{c|}{} & Openhands\_04\_15~\cite{wang2024openhands} & $68$ \\
\multicolumn{1}{c|}{} & \cellcolor[HTML]{C0C0C0}InfantAgent-Next + Claude-3.7-Sonnet & \cellcolor[HTML]{C0C0C0}$66$ \\
\multicolumn{1}{c|}{} & AutoCodeRover-v2.0 (Claude-3.5-Sonnet-20241022)~\cite{AutoCodeRover} & $52$ \\
\multicolumn{1}{c|}{} & SWE-agent + SWE-agent-LM-32B~\cite{SWE-agent} & $46$ \\
\multicolumn{1}{c|}{} & AppMap Navie v2~\cite{AppMap} & $12$ \\
\multicolumn{1}{c|}{\multirow{-7}{*}{Open Source}} & Agentless Lite + O3 Mini (20250214)~\cite{agentless} & $10$ \\
\hline
\multicolumn{1}{c|}{} & CodeStory Midwit Agent + swe-search~\cite{AgentFarm} & $70$ \\
\multicolumn{1}{c|}{} & AgentScope~\cite{agentscope} & $66$ \\
\multicolumn{1}{c|}{} & CORTEXA~\cite{CORTEXA} & $62$ \\
\multicolumn{1}{c|}{} & Amazon Q Developer Agent\_2024\_12\_02~\cite{AmazonQDeveloper} & $54$ \\
\multicolumn{1}{c|}{\multirow{-5}{*}{Close Source}} & devlo\_2024\_11\_08~\cite{devlo} & $48$ \\
\hline
\multicolumn{3}{c}{\textbf{SWE-Bench-Lite (using GPT-4o)}} \\
\hline
\multicolumn{1}{c|}{} & Agentless-1.5~\cite{agentless} & $32.00$ \\
\multicolumn{1}{c|}{} & \cellcolor[HTML]{C0C0C0}InfantAgent-Next & \cellcolor[HTML]{C0C0C0}$31.67$ \\
\multicolumn{1}{c|}{} & Agentless + RepoGraph~\cite{agentless} & $29.67$ \\
\multicolumn{1}{c|}{\multirow{-4}{*}{Open Source}} & OpenHands + CodeAct v1.8~\cite{wang2024openhands} & 22.00 \\
\hline
\multicolumn{1}{c|}{} & CodeShellTester~\cite{codeshell} & $31.33$ \\
\multicolumn{1}{c|}{\multirow{-2}{*}{Close Source}} & SIMA~\cite{SIMA} & $27.67$ \\
\hline
\end{tabular}
\end{adjustbox}
\vspace{-2pt}
\end{table}

\noindent{\textbf{\textit{Benchmark.}}} To rigorously evaluate the logical reasoning capabilities of \textsc{InfantAgent-Next}, we employ two complementary SWE-Bench~\cite{jimenez2024swebench} datasets. First, SWE-Bench-Lite, a 300-issue subset, challenges the agent to interpret real GitHub bug reports, identify defects in a Python codebase, propose patches, and verify fixes by executing the provided test suites. This setup emphasizes multi-step logical planning rather than GUI proficiency. Second, we assess performance on the full SWE-Bench-Verified suite (500 cases); to manage API-related costs, we uniformly sample 50 cases for evaluation (their instance IDs are listed in the Appendix~\ref{test_cases}). Together, these benchmarks provide a thorough yet cost-effective assessment of our agent’s reasoning capabilities.

\noindent{\textbf{\textit{Experiment Setup.}}} 
For SWE-Bench-Lite, our inference is conducted using GPT-4o for planning, tool selection, and execution, consistent with the configurations adopted by most existing agents to ensure fair comparisons. For the 50-case SWE-Bench-Verified subset, we employ \texttt{Claude-3.7-Sonnet} for each step, providing direct evaluation against state-of-the-art systems. In both settings, \texttt{DeepSeek-V3-0324} is used as the dedicated file-editing model.

\noindent{\textbf{\textit{Results.}}}
On the SWE-Bench-Verified benchmark (50 cases), \textsc{InfantAgent-Next} achieves a leading accuracy of \textbf{66\%}, outperforming many proprietary agents. Notably, several closed-source agents such as Amazon Q Developer Agent and Emergent E1 perform worse, highlighting the effectiveness of our agent's architecture and integration.
On the SWE-Bench-Lite benchmark, where we ensure a fair comparison by restricting all agents to use GPT-4o, \textsc{InfantAgent-Next} maintains competitive performance with an accuracy of \textbf{31.67\%}. This places our system among the top-performing open-source agents, demonstrating its robustness and adaptability.

\vspace{-3pt}
\subsection{General Task Performance}
\noindent{\textbf{\textit{Benchmark.}}}
We evaluate \textsc{InfantAgent-Next} on the GAIA benchmark~\cite{mialon2023gaiabenchmarkgeneralai}, designed to assess general AI assistants. GAIA comprises open-ended, real-world questions across three difficulty tiers—Level 1 (basic), Level 2 (intermediate), and Level 3 (advanced). These tasks require integrating core capabilities, including reasoning, multimodal understanding, web navigation, and tool usage, to produce a single, verifiable answer.

\noindent{\textbf{\textit{Experiment Setup.}}} 
The evaluation is conducted on the GAIA benchmark validation set. We employ \texttt{Claude-3.7-Sonnet} for reasoning and task execution, \texttt{Deepseek-V3-0324} for tool selection, \texttt{gpt-4o-audio-preview} for auditory processing, and \texttt{UI-TARS-1.5-7B}~\cite{qin2025ui} for visual understanding. System prompts are sourced directly from the official GAIA leaderboard to ensure consistency with standard evaluation protocols.

\noindent{\textbf{\textit{Results.}}}
The performance on the GAIA benchmark is presented in Table~\ref{tab:GAIA}. While closed-source agents maintain a performance edge, \textsc{InfantAgent-Next} ranks second among open-source agents, trailing only OWL, and achieves state-of-the-art results on Level 2 difficulty questions.

\begin{table}[]
\centering
\caption{Performance of \textsc{InfantAgent-Next} on GAIA. The data compared in this table are current as of the completion date of our GAIA experiments on April 17, 2025. For results without citations, please refer to the official GAIA leaderboard~\cite{gaia-benchmark_leaderboard_2025}.}
\label{tab:GAIA}
\begin{adjustbox}{max width=\textwidth}
\begin{tabular}{ccccccc}
\hline
Open/Close Source                                    & Agent Name                               & Main Model                                     & Average                       & Level 1                       & Level 2                       & Level 3                       \\ \hline
\multicolumn{1}{c|}{}                                & Langfun Agent v2.1~\cite{langfun}                       & Claude-3.7-sonnet                         & $71.52$                         & $83.02$                         & $68.60$                         & $57.69$                         \\
\multicolumn{1}{c|}{}                                & Trase Agent v0.3~\cite{TRASE}                         & o1                                        & $70.30$                         & $83.02$                         & $69.77$                         & $46.15$                         \\
\multicolumn{1}{c|}{}                                & h2oGPTe Agent v1.6.8~\cite{H2O.ai}                     & Claude-3.5-Sonnet                         & $63.64$                         & $67.92$                         & $67.44$                         & $42.31$                         \\
\multicolumn{1}{c|}{}                                & Anges                                    & -                                         & 60.00                         & 66.04                         & 65.12                         & 30.77                         \\
\multicolumn{1}{c|}{}                                & desearch                                 & GPT-4o                                    & $56.97$                         & $71.70$                         & $58.14$                         & $23.08$                         \\
\multicolumn{1}{c|}{}                                & Ormind v0.1~\cite{Ormind}                              & -                                         & $55.15$                         & $69.81$                         & $54.65$                         & $26.92$                         \\
\multicolumn{1}{c|}{}                                & Barcelona v0.1                           & Claude-3.5-sonnet                         & $50.30$                         & $62.26$                         & $50.00$                         & $26.92$                         \\
\multicolumn{1}{c|}{}                                & DRP-val-v.1.0                            & -                                         & $46.06$                         & $56.60$                         & $48.84$                         & $15.38$                         \\
\multicolumn{1}{c|}{}                                & omne                                     & o1-preview                                & $46.06$                         & $60.38$                         & $44.19$                         & $23.08$                         \\
\multicolumn{1}{c|}{}                                & qbnlp                                    & o1                                        & $44.24$                         & $52.83$                         & $45.35$                         & $23.08$                         \\
\multicolumn{1}{c|}{}                                & LRC-Huawei                               & -                                         & $40.61$                         & $52.83$                         & $43.02$                         & $7.69$                          \\
\multicolumn{1}{c|}{}                                & Agent-On-the-fly                         & QwQ-32B                                   & $40.61$                         & $52.83$                         & $43.02$                         & $7.69$                          \\
\multicolumn{1}{c|}{\multirow{-13}{*}{Close Source}} & AgentIM v1.1                             & gpt-4-turbo                               & $40.00$                         & $50.94$                         & $40.70$                         & $15.38$                         \\ \hline
\multicolumn{1}{c|}{}                                & OWL-Roleplaying                          & GPT-4o + o3-mini                          & $\mathbf{58.18}$                         & $\mathbf{81.13}$                         & $54.65$                         & $23.08$                         \\
\multicolumn{1}{c|}{}                                & \cellcolor[HTML]{C0C0C0}InfantAgent-Next & \cellcolor[HTML]{C0C0C0}Claude-3.7-Sonnet & \cellcolor[HTML]{C0C0C0}$56.97$ & \cellcolor[HTML]{C0C0C0}$62.26$ & \cellcolor[HTML]{C0C0C0}$\mathbf{62.79}$ & \cellcolor[HTML]{C0C0C0}$26.92$ \\
\multicolumn{1}{c|}{}                                & TapeAgents~\cite{TapeAgents}                               & Claude-3.7-Sonnet                         & $55.76$                         & $71.70$                        & $53.49$                         & $\mathbf{30.77}$                         \\
\multicolumn{1}{c|}{}                                & Auto-Deep-Research~\cite{Auto-Deep-Research}                       & Claude-3.7-Sonnet                         & $55.15$                         & $71.70$                         & $53.49$                         & $26.92$                         \\
\multicolumn{1}{c|}{}                                & AutoAgent                                & -                                         & $55.15$                         & $71.70$                         & $53.49$                         & $30.77$                         \\
\multicolumn{1}{c|}{}                                & Open Deep Research~\cite{open-deep-research}                       & o1                                        & $55.15$                         & $67.92$                         & $53.49$                         & $34.62$                         \\
\multicolumn{1}{c|}{}                                & Magnetic-1                               & o1                                        & $46.06$                         & $56.60$                         & $46.51$                         & $23.08$                         \\
\multicolumn{1}{c|}{}                                & HuggingFace Agents~\cite{HuggingFaceAgents}                       & gpt-4o                                    & $44.24$                         & $58.49$                         & $43.02$                         & $19.23$                         \\
\multicolumn{1}{c|}{}                                & Multi-Agent Exp v0.1                     & gpt-4-turbo                               & $39.39$                         & $54.72$                         & $38.37$                         & $11.54$                         \\
\multicolumn{1}{c|}{\multirow{-10}{*}{Open Source}}  & FRIDAY~\cite{wu2024copilot}                                   & gpt-4-turbo                               & $34.55$                         & $45.28$                         & $34.88$                         & $11.54$                         \\ \hline
\end{tabular}
\end{adjustbox}
\vspace{-10pt}
\end{table}

\vspace{-1pt}
\subsection{Ablation on Visual Grounding design}
\noindent{\textbf{\textit{Benchmark.}}} ScreenSpot‑Pro~\cite{li2025screenspot} 
provides a high-resolution professional scenarios to stress‑test visual grounding. It contains 1,581 annotated screenshots drawn from 23 applications spanning five industry domains and three operating systems; each sample pairs a natural‑language prompt with the exact on‑screen target element, enabling precise click‑accuracy evaluation. 
We therefore use ScreenSpot‑Pro to quantify the performance of our newly designed Iterative Region Cropping.

\begin{figure}[htbp]
\centering
\begin{subfigure}[b]{0.24\textwidth}
    \includegraphics[width=\textwidth]{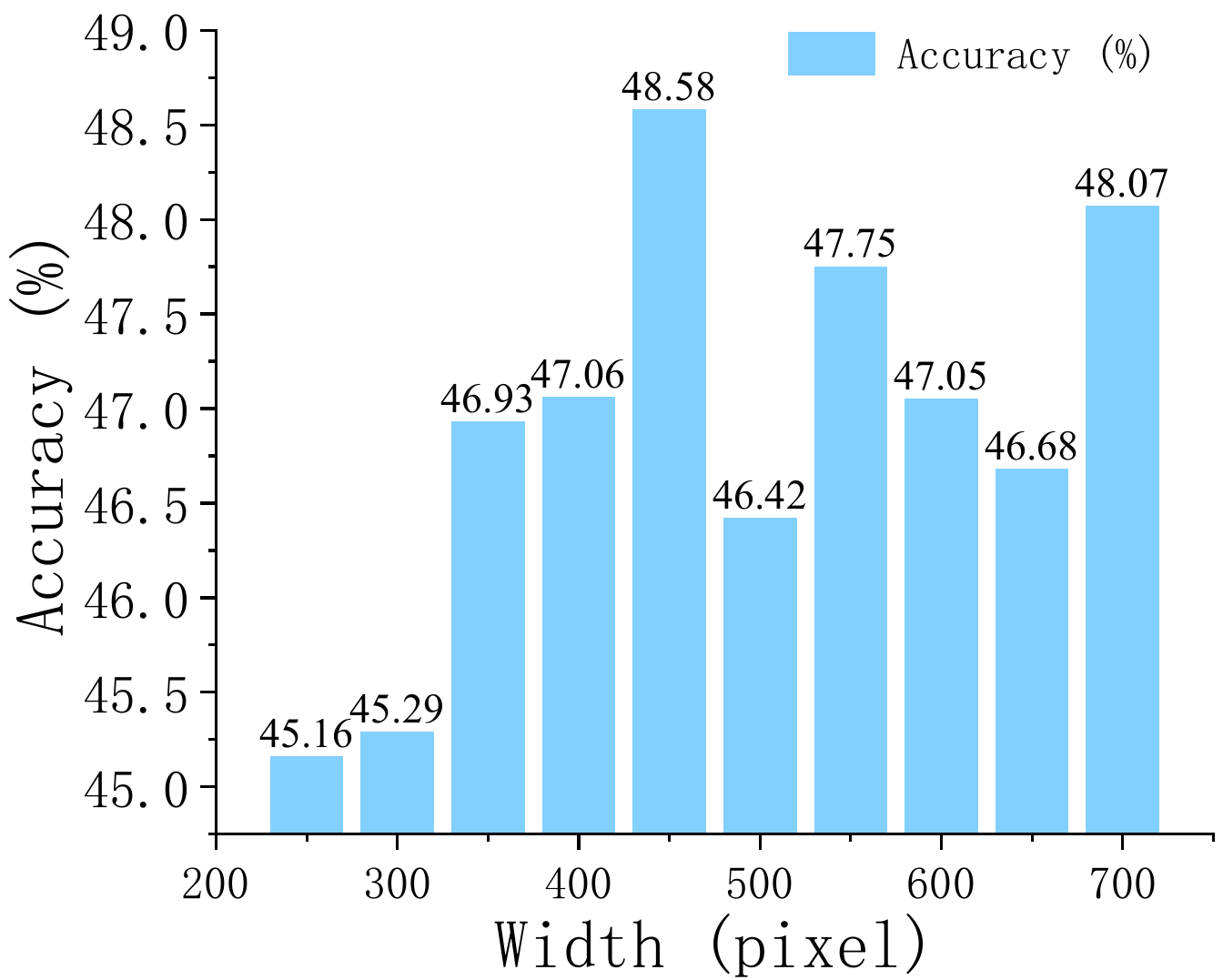}
    \caption{}
    \label{fig:image-a}
\end{subfigure}
\hfill
\begin{subfigure}[b]{0.24\textwidth}
    \includegraphics[width=\textwidth]{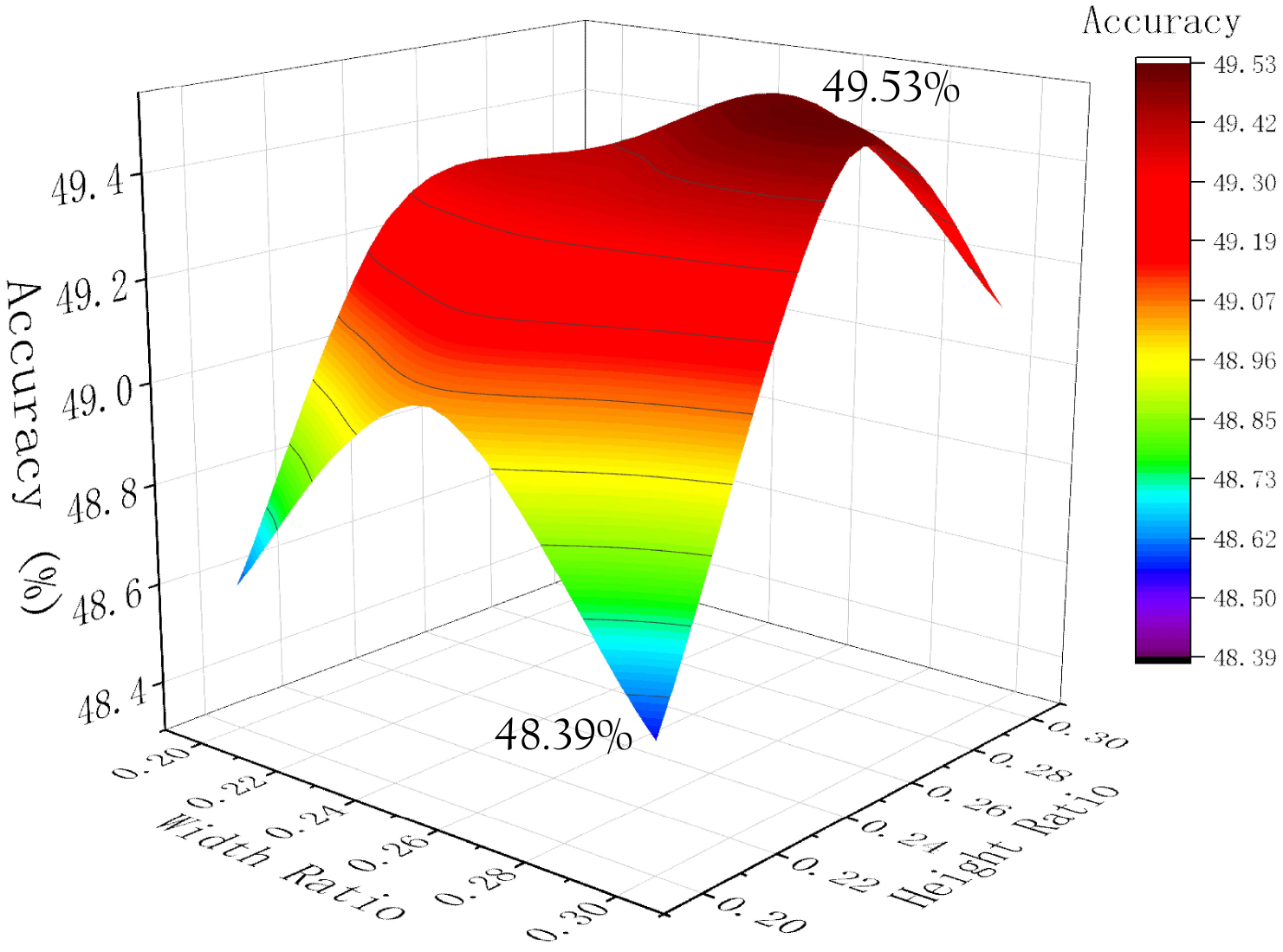}
    \caption{}
    \label{fig:image-b}
\end{subfigure}
\hfill
\begin{subfigure}[b]{0.24\textwidth}
    \includegraphics[width=\textwidth]{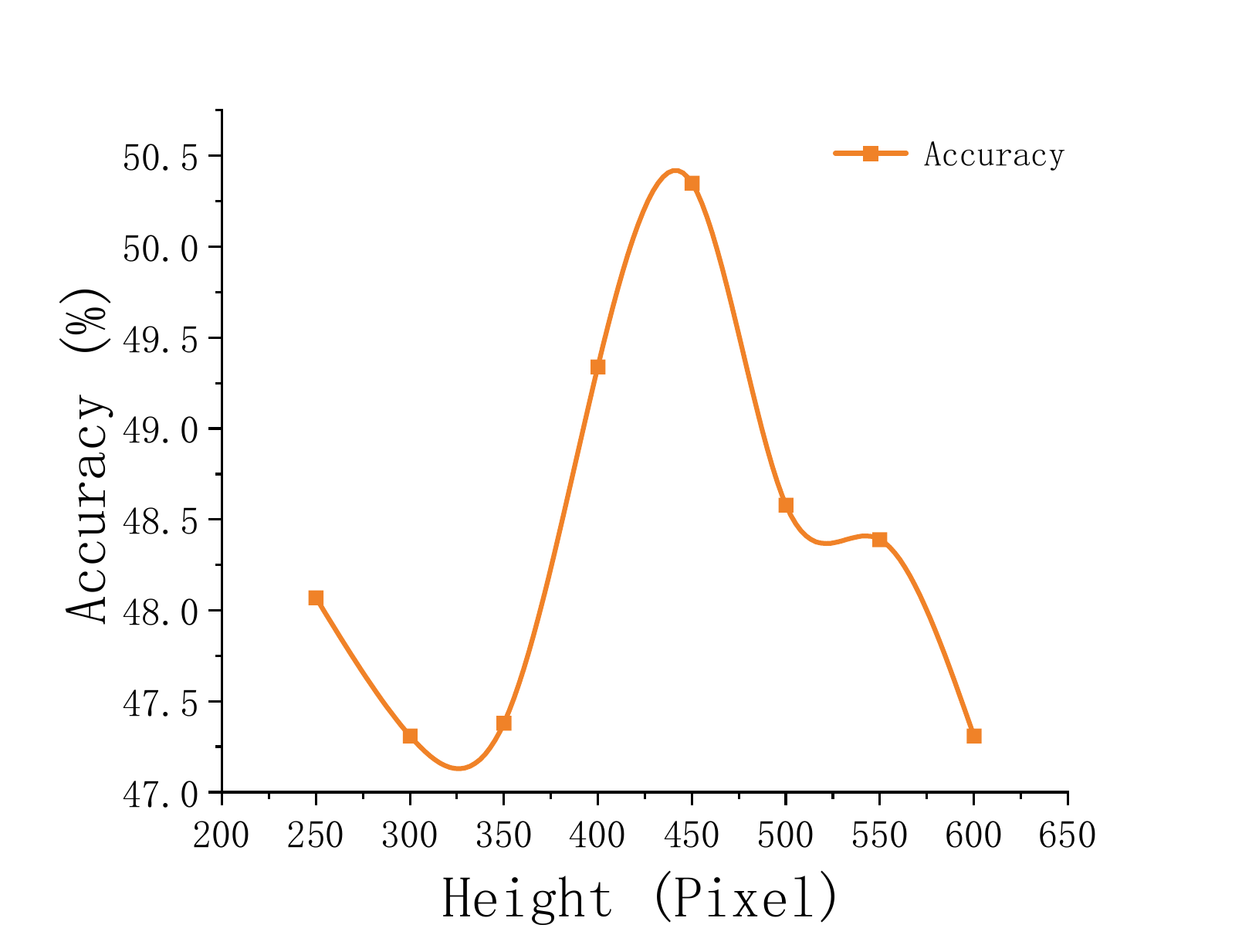}
    \caption{}
    \label{fig:image-c}
\end{subfigure}
\hfill
\begin{subfigure}[b]{0.24\textwidth}
    \includegraphics[width=\textwidth]{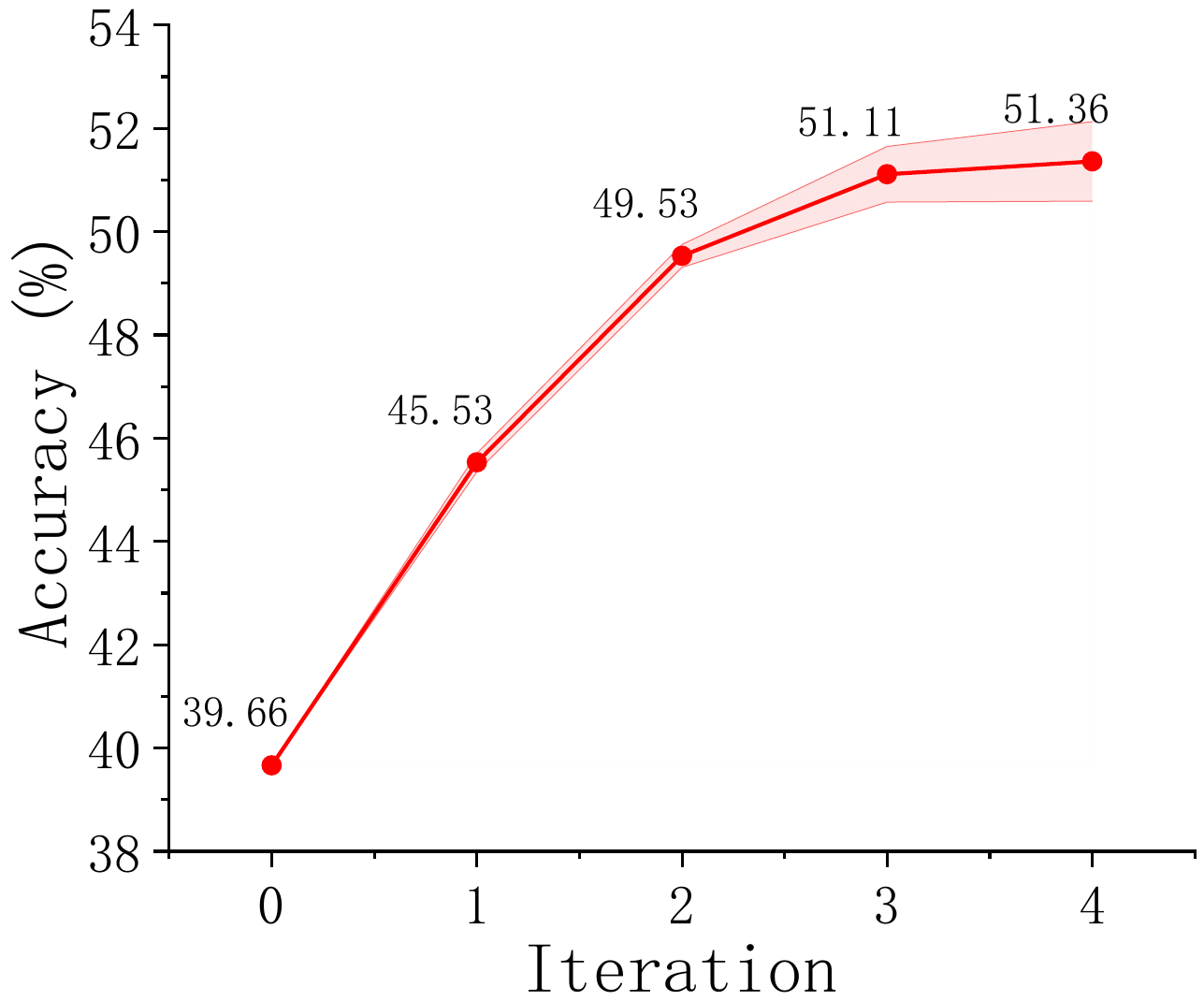}
    \caption{}
    \label{fig:image-d}
\end{subfigure}
\caption{We conduct an ablation study on the Iterative Region Cropping setup from four perspectives.
(a) Vary the region width.
(b) Vary the width and height ratios of the cropped region (relative to the full image)
(c) Vary the height.
(d) Vary the number of iteration}
\label{fig:four-images}
\end{figure}

\noindent{\textbf{\textit{Experiment Setup.}}}
We performed our experiment using 2x A100 80G GPU.
We first conduct an ablation study on the Iterative Region Cropping setup.
Specifically, for \texttt{CROP\_SIZE}, we test three different settings with a fixed iteration length of $n=2$ to evaluate its impact:
(Fig.~\ref{fig:four-images} (a)) we fix the cropped region area to 150K pixels and vary the region width.
(Fig.~\ref{fig:four-images} (b)) we examine how varying the width and height ratios of the cropped region (relative to the full image) affects performance.
(Fig.~\ref{fig:four-images} (c)) we fix the width at 700 pixels and vary the height. 
(Fig.~\ref{fig:four-images} (d)) we further ablate the number of iterations, fixing the region width and height ratios to 0.3 and 0.25, respectively. 

\noindent\textbf{\textit{Results.}} 
Among all strategies, 
the setting in Fig.\ref{fig:four-images} (c)—fixing the width while varying the height—yields the highest accuracy and is preferred for determining \texttt{CROP\_SIZE}.
For Fig.~\ref{fig:four-images} (d), accuracy converges at the third iteration, and using two iterations offers a favorable trade-off between performance and inference cost.

\vspace{-2pt}
\subsection{Evaluation on File Editing}
\noindent{\textbf{\textit{Benchmark.}}} 
We evaluate the file editing capabilities of \textsc{InfantAgent-Next} using the SWE-Bench-Verified dataset~\cite{jimenez2024swebench}, a curated subset of 500 real-world GitHub Issue–Pull Request pairs from popular Python projects. Each task is human-validated to ensure sufficient context and is verifiable through the project’s unit tests. Tasks involve diverse code modification instructions, such as deleting or updating code across multiple files, testing the agent’s ability to interpret and implement complex changes. We randomly sampled 10\% of the tasks from the SWE-Bench-Verified dataset across repository categories to form our evaluation set. The IDs of the selected test cases are in the Appendix~\ref{test_cases}.

\noindent{\textbf{\textit{Experiment Setup.}}} 
We evaluate each task solution and categorize them into three sets:  
$C_{\mathrm{total}}$, the set of all task solutions;  
$C_{\mathrm{fail}}$, solutions failing due to execution-model errors;  
$C_{\mathrm{repaired}}$, solutions successfully repaired.  
We define the following metrics:
$\mathrm{RepairSuccessRate~(RSR)} =\lvert C_{\mathrm{repaired}}\rvert/\lvert C_{\mathrm{fail}}\rvert$ and $\mathrm{OverallRepairRate ~(ORR)} ={\lvert C_{\mathrm{repaired}}\rvert}/{\lvert C_{\mathrm{total}}\rvert}$

\begin{wrapfigure}{r}{0.65\textwidth}
\vspace{-8pt}
  \centering
  \includegraphics[width=0.65\textwidth]{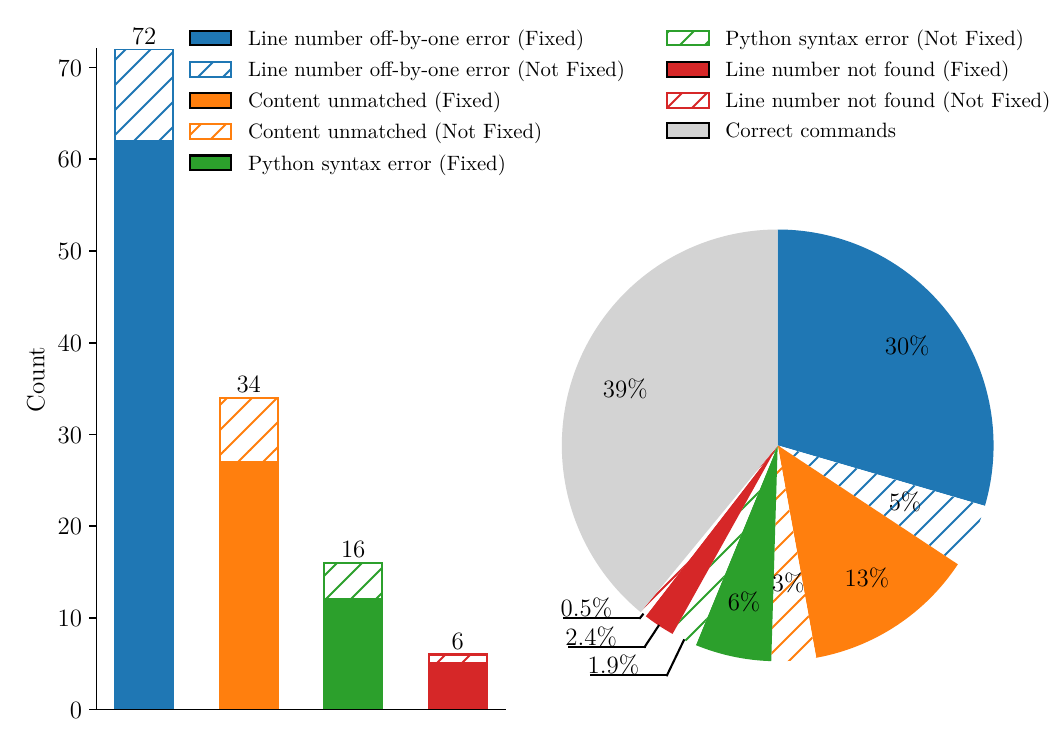}
    \caption{Evaluation on a subset of SWE-Bench-Verified.}
  \label{fig:tool_shape}
\end{wrapfigure}
\noindent\textbf{\textit{Results.}} \textsc{InfantAgent-Next} exhibits strong performance on file editing tasks in the SWE-Bench-Verified dataset, achieving a total success rate of 90.4\% with 84.3\% RSR and 51.4\% ORR.
The agent excels at fixing line number off-by-one errors (30\% repaired), content unmatched (13\% repaired), and Python syntax errors (6\% repaired), demonstrating robust code interpretation and repair capabilities. Overall, these results highlight \textsc{InfantAgent-Next} as a highly capable agent for automated file editing, with minor limitations in edge cases.

%% file: Sections/Conclusion.tex
\section{Conclusion}

We presented \textsc{InfantAgent-Next}, a multimodal generalist agent that bridges the strengths of tool-based and pure vision-based paradigms through a modular, context-aware architecture. By routing subtasks to specialized models and maintaining a unified dialogue context, \textsc{InfantAgent-Next} avoids the limitations of single-model systems and delivers both high task accuracy and broad applicability across diverse interfaces. Empirical results on OSWorld, GAIA, and SWE-Bench confirm its effectiveness, including a $\mathbf{7.27\%}$ accuracy gain over Claude-Computer-Use. All code, models, and evaluation tools will be released to support future research in multimodal agent design.

%% file: Sections/Acknowledgements.tex
\section*{Acknowledgment}




This research was partially supported by Cisco Research. We thank the open-source community for their valuable suggestions, issues, and pull requests.

%% file: Sections/Appendix.tex
\newpage
\appendix
\section{Case Analysis}
\label{apd:case_analysis}





\begin{figure}[ht]
  \centering
  \includegraphics[width=0.99\textwidth]{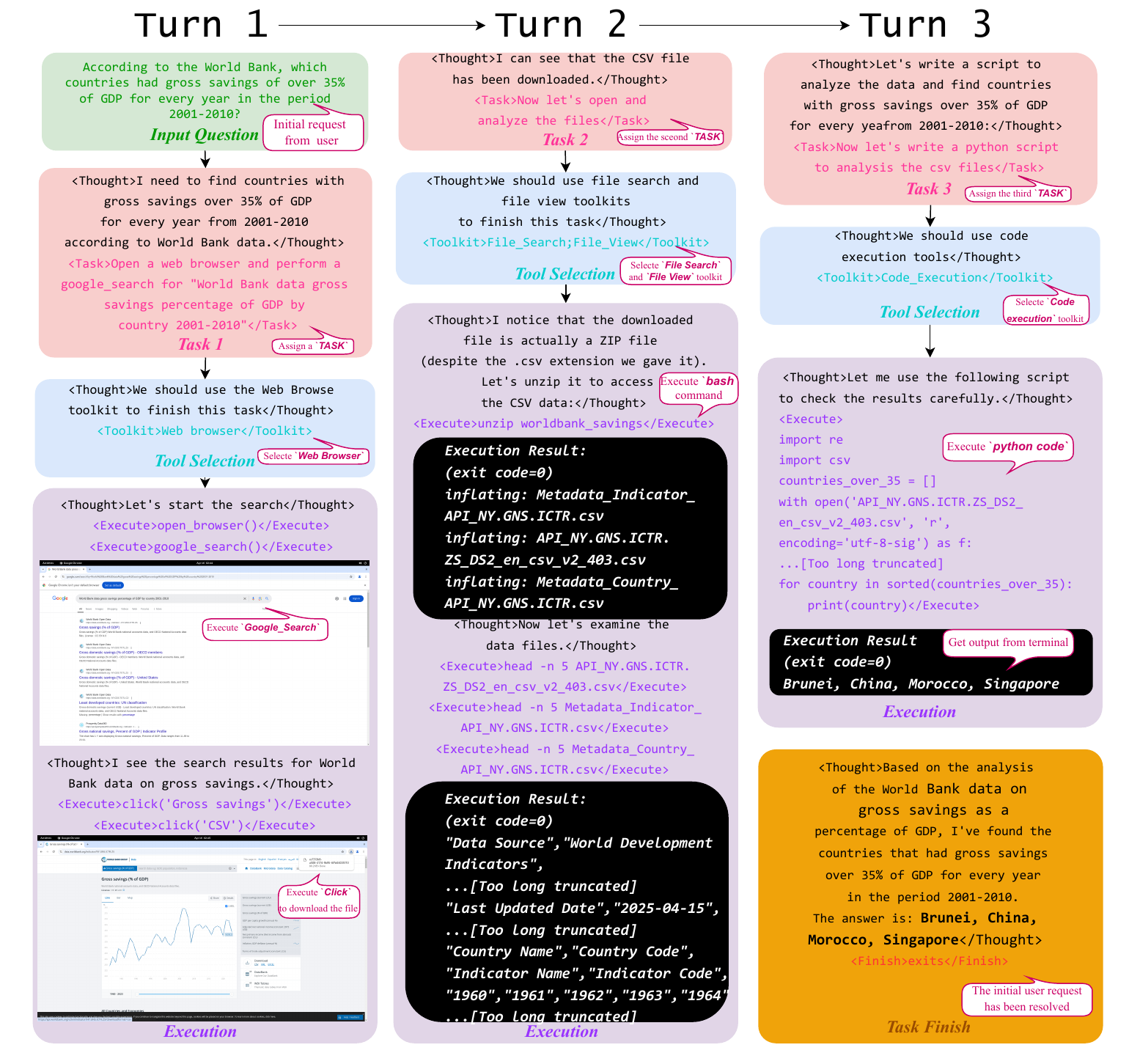} 
\caption{Cases analysis. Zoom in to view the detailed content in the screenshot.}
  \label{fig:case_analysis}
\end{figure}

Figure~\ref{fig:case_analysis} illustrates the step-by-step process by which \textsc{InfantAgent-Next} solves a real-world query:
\textit{“According to the World Bank, which countries had gross savings of over 35\% of GDP for every year in the period 2001–2010?”}

To answer this question, \textsc{InfantAgent-Next} performs three sequential cycles of \textit{Planning} → \textit{Tool Selection} → \textit{Execution}:

\textbf{First loop:} The agent selects the \texttt{Web\_Browser} toolkit, conducts a web search, performs mouse-click operations, and captures desktop screenshots as execution outputs. This enables it to successfully locate and download the relevant CSV file.

\textbf{Second loop:} It then selects the \texttt{File\_Search} and \texttt{File\_View} toolkits, using Bash commands to inspect the directory structure and examine the contents of the CSV file.

\textbf{Third loop:} Finally, the agent selects the \texttt{Code\_Execution} toolkit and executes Python code to parse and aggregate the data in order to identify qualifying countries.

Through these modular and interpretable steps, \textsc{InfantAgent-Next} arrives at the correct answer with minimal human intervention.

\section{Workflow Prompt Templates}
\subsection{Planning}
\label{apd:planning_prompt_template}
\begin{lstlisting}[caption={Planning system prompt. It is inserted at the beginning of each planning.}]
In our interaction with the **User (requester)**, our goal is to gradually resolve their request or answer their questions. 
The process involves three roles: **You (reasoner)**, **Me (executor)**, and **User (requester)**.
In each response, you must execute only one step, choosing to either assign a task to **Me (executor)**, or request information from the **User (requester)**.
**I (executor)** can help **You (reasoner)** complete tasks that **You (reasoner)** cannot perform directly, including:
{task_category}
Our goal is to resolve the **User (requester)**'s request step by step.
* If **You (reasoner)** want to assign a task to **Me (executor)**, please use the <task>...</task> tag to describe your task, don't give the detail execution commands for now.
* If **You (reasoner)** want to request information from the **User (requester)**, ask them directly without using any tags.
If you believe the user's request has already been resolved, please answer the **User (requester)**'s request based on the entire conversation history and at the end of your answer add <finish>exit</finish>.
Here is an example:
USER: <user_request>
Can you help me to download a PDF file called example.pdf from the internet and convert its Table 1 to .csv file?
</user_request>
ASSISTANT: <task>Download the example.pdf from the internet.</task>
USER: [Downloads output]
ASSISTANT: <task>Convert Table 1 of example.pdf to .csv file.</task>
USER: [Converts output]
ASSISTANT: <task>Verify if the .csv file exist.</task>
USER: [Terminal output]
ASSISTANT: Now the task is finished and we can provide the .csv file to the user. <finish>exit</finish>
\end{lstlisting}

\begin{lstlisting}[caption={Planning end prompt. It is appended to the end of each planning.}]
If you believe the user's request has already been resolved, please answer the **User (requester)**'s request based on the entire conversation history and at the end of your answer add <finish>exit</finish>.
Otherwise, provide assign a task to **Me (executor)** within the appropriate execution tag:
> Use the <task>...</task> tag for the task that you would like to assign to me.
Please do NOT repeat the similar analysis that you have already provided.
\end{lstlisting}

\begin{lstlisting}[caption={Planning avoid repetition prompt. It is appended to the end of the planning dialogue whenever the planning process is needlessly repeated.}]
If you believe the user's request has already been resolved, please answer the **User (requester)**'s request based on the entire conversation history and at the end of your answer add <finish>exit</finish>.
Otherwise, please assign a task based on your analysis. 
Please do NOT repeat the similar analysis/Task again!
\end{lstlisting}

\subsection{Tool Selection}
\label{apd:tool_selection_prompt_template}
\begin{lstlisting}[caption={Tool Selection System Prompt.  it is inserted at the beginning of each tool selection action.}]
I would like to finish a task, please help me choose the suitable sets of commands to complete this task.
1. File editing related commands: 
This set of commands can be used to view file content, as well as perform additions, deletions, searches, and modifications on files. 
If you want to select this set of commands, please return: <toolkit>file_edit</toolkit>
2. Code execution related commands: 
This set of commands can be used to execute code snippets. 
If you want to select this set of commands, please return: <toolkit>code_exec</toolkit>
3. Computer interaction commands: 
These commands can be used to interact with the computer via the keyboard and mouse. 
If you want to select this set of commands, please return: <toolkit>computer_interaction</toolkit>
4. Web browsing related commands: 
This set of commands can be used to interact with web pages. 
If you want to select this set of commands, please return: <toolkit>web_browse</toolkit>
5. File understanding related commands: 
This set of commands can be used to understand the content of files. Such as reading files, view images, listen to audios, watch videos, etc.
If you want to select this set of commands, please return: <toolkit>file_understand</toolkit>
If you want to select multiple sets of commands, please separate them with commas. 
For example, if you think we not only need to edit some files but also execute some code, you should return: <toolkit>file_edit, code_exec</toolkit>.
\end{lstlisting}

\subsection{Execution}
\label{apd:execution_prompt_template}
\begin{lstlisting}[caption={Execution system prompt. it is inserted at the beginning of execution. It includes usage instructions and examples for the selected tool functions.}]
{Selected_Tool_Guide_Book}
{Selected_Tool_Examples}
\end{lstlisting}

\begin{lstlisting}[caption={Execution end prompt. it is appended to the end of the execution.}]
If you think the current task: {task} is already solved, please respond with your conclusion and include the following tag at the end:
<task_finish>
exit
</task_finish>. 
Otherwise, provide the next command within the appropriate execution tag:
> Use <execute_bash>...</execute_bash> for Bash commands.
> Use <execute_python>...</execute_python> for my other customized commands, as I mentioned in the beginning.
Please don't take any steps on your own that aren NOT related to completing the current task: {task}; I will guide you through the next steps. 
You only need to ensure that the current task is completed.
\end{lstlisting}

\section{Toolkits}
\label{apd:Toolkits}
\subsection{File Edit Toolkit}
\begin{lstlisting}[caption={File edit tool functions}]
Please use the following file editing functions to add, delete, search, and modify files.
- create_file(filename: str, content: str | None): Creates and opens a new file with the given name. Add the content to the file if content is not None.
- replace_content(file_path, old_content, new_content): Replaces the old content with the new content in the specified file. For the old_content/new_content argument, please focus only on the parts that actually need to be changed.
- edit_file(file_name: str, start_line: int, start_str: str, end_line: int, end_str: str, content: str): Edits the specified file by replacing the content between start and end lines with the new content. file_name: Name of the file. start_line: Starting line number. start_str: String content in Starting line. end_line: Ending line number. end_str: String content in Ending line. content: New content to replace.
- append_file(file_name, content, start_line): Appends given content to a file. file_name: Name of the file. content: Content to append. start_line: Line number to start appending from (default is the end of the file).
- search_function(file_path, function_signature): Search and show a function in the file. For the function_signature, you should only input the function name.
\end{lstlisting}

\subsection{File View Toolkit}
\begin{lstlisting}[caption={File view tool functions}]
Please use the following functions to understand the content of files. Such as reading files, view images, listen to audios, watch videos, etc.
- open_file(path: str, line_number: int | None = 1, context_lines: int = 100): Opens a file (txt, csv, word, code file, etc.) and optionally moves to a specific line. path: Path to the file. line_number: Line number to move to. context_lines: Number of lines to display.
- parse_pdf(pdf_path: str, page: int): View the specified page of a PDF file. pdf_path: Path to the PDF file. page: Page number to view.
- parse_figure(figure_path: str): View the specified figure. figure_path: Path to the figure file.
- parse_audio(audio_path: str, question: str): Ask a question about the audio file. audio_path: Path to the audio file. question: Question to ask.
- zoom_pdf(pdf_path: str, page: int, region: tuple): Zoom in on a specific region of a PDF file. pdf_path: Path to the PDF file. page: Page number to view. region: Tuple specifying the region to zoom in on (x0, y0, x1, y1).
\end{lstlisting}

\subsection{File Search Toolkit}
\begin{lstlisting}[caption={File search tool functions}]
Please use the following functions to search files.
- search_dir(search_term, dir_path='./'): Searches for a term in all files in the specified directory.
- find_file(file_name, dir_path='./'): Finds all files with the given name in the specified directory.
\end{lstlisting}

\subsection{Web Browser Toolkit}
\begin{lstlisting}[caption={Web browser tool functions}]
You can use the following functions to interact with the browser.
- open_browser(): Open the browser.
- navigate_to(url: str) : Navigate to the specified URL.
- refresh_page(): Refresh the current page.
- go_back(): Go back to the previous page.
- go_forward(): Go forward to the next page.
- close_current_tab(): Close the current tab.
- execute_javascript(script: str): Execute the specified JavaScript code.
- switch_to_tab(page_id: int): Switch to the tab at the specified index.
- create_new_tab(url: str): Open a new tab with the specified URL.
- save_cookies(): Save the current cookies.
- select_dropdown_option(selector_index: int, option: int): Select the specified option from the dropdown menu. selector_index: selector index. option: Index of the option to select.
- google_search(content: str): Perform a Google search for the specified content. content: The content to search for.
- close(): Close the browser.
\end{lstlisting}

\subsection{Computer Use Toolkit}
\begin{lstlisting}[caption={Computer use functions}]
You can use the following functions to perform various mouse and keyboard operations.
- clear_text(): Clear the text in the current input field. please make sure the input field is selected before using this command.
- take_screenshot(): If you want to check the current screen, you can use this command to take a screenshot of the current screen.
- mouse_left_click(item: str, description: str): Left mouse click at the specified position. For example: mouse_left_click('search bar', 'It is located near the top center of the Google Chrome browser window. It is a long, rectangular input field with rounded edges. The search bar spans almost the entire width of the browser window and sits directly below the browser's tab row. It has placeholder text that reads "Search Google or type a URL." The search bar is centrally aligned, making it easy to spot above the main content area of the browser.')
- mouse_double_click(item: str, description: str): Double-click at the specified position. For example: mouse_double_click('The VSCode icon', 'It is located in the sidebar (Launcher) on the left side of the screen. It is the first icon from the top in the vertical arrangement. The icon has a blue background with a white folded "V"-shaped design in the center. The sidebar is aligned along the leftmost edge of the screen, adjacent to the desktop background on its right side.')
- mouse_right_click(item: str, description: str): Right mouse click at the specified position. For example: mouse_right_click('The refresh button', 'It is located at the top-left corner of the Google Chrome browser window, inside the address bar. It is a circular arrow icon situated next to the left and right navigation arrows (back and forward buttons). The refresh button is just to the left of the search bar. Click it to refresh the current page.')
- mouse_scroll(direction: str, amount: int): Scroll mouse scroll up or down. direction: Direction to scroll ("up" or "down"). amount: Number of times to scroll.
- type_text(text: str): Type the given text. text: The text to type.
- download(url: str, save_dir: str): If you know the url of the file, you can use this command to download the file from the specified URL to the specified directory.
- press_key(key: str): Presses the specified key. key: The key or key combination to press (e.g., "Return", "Ctrl+c").
- open_application(app_name: str): Opens a specific application using the system application launcher. app_name: The name of the application to open (e.g., "chrome").
- mouse_drag(x_start: int, y_start: int, x_end: int, y_end: int): Drag the mouse from one position to another. x_start: Starting x-coordinate. y_start: Starting y-coordinate. x_end: Ending x-coordinate. y_end: Ending y-coordinate.
- mouse_box_select(x_start: int, y_start: int, x_end: int, y_end: int): Selects a box by dragging the mouse from one position to another. x_start: Starting x-coordinate. y_start: Starting y-coordinate. x_end: Ending x-coordinate. y_end: Ending y-coordinate.
\end{lstlisting}

\subsection{Code Execution Toolkit}
\begin{lstlisting}[caption={Code execution tool}]
If no suitable command is available, you can also use bash commands to interact with the terminal.
Do not use bash commands to edit/create the files. Instead, use the file editing functions provided.
\end{lstlisting}

\subsection{Advanced Toolkit}
\begin{lstlisting}[caption={Advanced tools}]
You can use the following functions to perform advanced operations. These commands are compound commands. When available, please prefer using these commands rather than attempting to perform the operations by yourself.
- search_arxiv(keyword: str, start_date: str, end_date: str, subject: str, field: str) : Searches for papers on arXiv based on the given keyword, date range, subject (options: cs, math, physics, q-bio, q-fin, stat), and keyword field (options: title, abstract, comments, author, all). Returns the search results.
- download_arxiv_pdf(arxiv_id: str) : Downloads the specified arXiv paper based on its ID (eg: 1608.06816) and show the first page of the PDF.
- scroll_pdf_page(direction: str, pages: int): When you're viewing a PDF document in the web interface, you can use this command to scroll the specified page of the PDF file up or down. direction: Direction to scroll ("up" or "down"). page: Page number to scroll.
- watch_video(video_path_or_url: str): Please use this command to Watch a video file or YouTube URL. Especially when there is a sign-in prompt for Youtube. video_path_or_url: Local path or YouTube URL.
- count_string_in_pdf(pdf_path: str, search_string: str): Count the number of occurrences of a specific string in a **Local** PDF file. pdf_path: Path to the PDF file. search_string: The string to search for.
.
\end{lstlisting}

\section{Details of Dedicated Computer}\label{Dedicated}
A fully isolated execution environment enhances security and enables full-spectrum command execution. Many existing agents (e.g., AutoGPT~\cite{Significant-Gravitas_AutoGPT_2025}) run directly on the host machine or run CLI operations within Docker containers. In contrast, our \textsc{InfantAgent-Next} provides a dedicated, standalone computing environment that supports command-line interaction via Bash, Python scripts via Jupyter, and direct GUI operations through the GNOME desktop.
Specifically, to ensure safety and control, the agent operates within an isolated virtual machine (VM), preventing direct access to the user’s computer and thus eliminating the risk of modifying or deleting critical files. At the same time, the user can still interact with the VM. The setup consists of three stages:
(1) Build a base Docker image;
(2) Enable GPU-accelerated rendering to support high frame rates and image quality;
(3) Expose the VM’s GNOME desktop via a remote desktop interface.
When the agent requires visual feedback (e.g., to perform a mouse click), the VM captures a screenshot of the current display and sends it back to the agent. The agent then analyzes the image before issuing the next command.

\section{Test Cases of SWE-Bench-Verified}\label{test_cases}

We detail the test cases of SWE-Bench-Verified in the following:

\begin{itemize}
  \item \texttt{astropy\_\_astropy-12907}
  \item \texttt{astropy\_\_astropy-14995}
  \item \texttt{astropy\_\_astropy-7606}
  \item \texttt{astropy\_\_astropy-8707}
  \item \texttt{django\_\_django-11451}
  \item \texttt{django\_\_django-11603}
  \item \texttt{django\_\_django-12858}
  \item \texttt{django\_\_django-13417}
  \item \texttt{django\_\_django-14500}
  \item \texttt{django\_\_django-15930}
  \item \texttt{django\_\_django-16032}
  \item \texttt{django\_\_django-16256}
  \item \texttt{django\_\_django-16899}
  \item \texttt{matplotlib\_\_matplotlib-20859}
  \item \texttt{matplotlib\_\_matplotlib-22719}
  \item \texttt{matplotlib\_\_matplotlib-24970}
  \item \texttt{matplotlib\_\_matplotlib-25122}
  \item \texttt{mwaskom\_\_seaborn-3069}
  \item \texttt{mwaskom\_\_seaborn-3187}
  \item \texttt{pallets\_\_flask-5014}
  \item \texttt{psf\_\_requests-1142}
  \item \texttt{psf\_\_requests-1766}
  \item \texttt{psf\_\_requests-1921}
  \item \texttt{psf\_\_requests-5414}
  \item \texttt{pydata\_\_xarray-4075}
  \item \texttt{pydata\_\_xarray-6599}
  \item \texttt{pydata\_\_xarray-6744}
  \item \texttt{pydata\_\_xarray-6938}
  \item \texttt{pylint-dev\_\_pylint-4661}
  \item \texttt{pylint-dev\_\_pylint-4970}
  \item \texttt{pylint-dev\_\_pylint-6386}
  \item \texttt{pylint-dev\_\_pylint-6528}
  \item \texttt{pytest-dev\_\_pytest-10081}
  \item \texttt{pytest-dev\_\_pytest-5631}
  \item \texttt{pytest-dev\_\_pytest-5809}
  \item \texttt{pytest-dev\_\_pytest-7205}
  \item \texttt{pytest-dev\_\_pytest-7432}
  \item \texttt{scikit-learn\_\_scikit-learn-10297}
  \item \texttt{scikit-learn\_\_scikit-learn-12585}
  \item \texttt{scikit-learn\_\_scikit-learn-12973}
  \item \texttt{scikit-learn\_\_scikit-learn-13135}
  \item \texttt{sphinx-doc\_\_sphinx-11445}
  \item \texttt{sphinx-doc\_\_sphinx-7454}
  \item \texttt{sphinx-doc\_\_sphinx-8035}
  \item \texttt{sphinx-doc\_\_sphinx-8551}
  \item \texttt{sphinx-doc\_\_sphinx-8721}
  \item \texttt{sympy\_\_sympy-13798}
  \item \texttt{sympy\_\_sympy-14531}
  \item \texttt{sympy\_\_sympy-16792}
  \item \texttt{sympy\_\_sympy-17630}
\end{itemize}

\section{Limitations and Further Work}\label{Limitations}
Our current work is confined to the reasoning phase. To mitigate the impact of excessive prompt engineering, we will next further train the model to automatically invoke the appropriate tools rather than relying on manually added prompts.
\newpage
\section*{NeurIPS Paper Checklist}

\begin{enumerate}

\item {\bf Claims}
    \item[] Question: Do the main claims made in the abstract and introduction accurately reflect the paper's contributions and scope?
    \item[] Answer: \answerYes{} 
    \item[] Justification: The abstract clearly state the claims made, including the contributions made in the paper.

\item {\bf Limitations}
    \item[] Question: Does the paper discuss the limitations of the work performed by the authors?
    \item[] Answer: \answerYes{} 
    \item[] Justification: Limitation is included in the Appendix.

\item {\bf Theory assumptions and proofs}
    \item[] Question: For each theoretical result, does the paper provide the full set of assumptions and a complete (and correct) proof?
    \item[] Answer: \answerNA{} 
    \item[] Justification: This paper is an experiment-driven study.

    \item {\bf Experimental result reproducibility}
    \item[] Question: Does the paper fully disclose all the information needed to reproduce the main experimental results of the paper to the extent that it affects the main claims and/or conclusions of the paper (regardless of whether the code and data are provided or not)?
    \item[] Answer: \answerYes{} 
    \item[] Justification: All the code and test scripts are included in the Supplemental Material.

\item {\bf Open access to data and code}
    \item[] Question: Does the paper provide open access to the data and code, with sufficient instructions to faithfully reproduce the main experimental results, as described in supplemental material?
    \item[] Answer: \answerYes{} 
    \item[] Justification: All the code and test scripts are included in the Supplemental Material.

\item {\bf Experimental setting/details}
    \item[] Question: Does the paper specify all the training and test details (e.g., data splits, hyperparameters, how they were chosen, type of optimizer, etc.) necessary to understand the results?
    \item[] Answer: \answerYes{} 
    \item[] Justification: We have the experiment setup section for each experiment.

\item {\bf Experiment statistical significance}
    \item[] Question: Does the paper report error bars suitably and correctly defined or other appropriate information about the statistical significance of the experiments?
    \item[] Answer: \answerYes{} 
    \item[] Justification: We have the error bar for Visual Grounding experiment.

\item {\bf Experiments compute resources}
    \item[] Question: For each experiment, does the paper provide sufficient information on the computer resources (type of compute workers, memory, time of execution) needed to reproduce the experiments?
    \item[] Answer: \answerYes{}, 
    \item[] Justification: We mentioned related information in our Visual Grounding experiment.
    
\item {\bf Code of ethics}
    \item[] Question: Does the research conducted in the paper conform, in every respect, with the NeurIPS Code of Ethics \url{https://neurips.cc/public/EthicsGuidelines}?
    \item[] Answer: \answerYes{} 
    \item[] Justification: We followed the NeurIPS Code of Ethics.

\item {\bf Broader impacts}
    \item[] Question: Does the paper discuss both potential positive societal impacts and negative societal impacts of the work performed?
    \item[] Answer: \answerYes{} 
    \item[] Justification: There is no societal impact of the work performed.
    
\item {\bf Safeguards}
    \item[] Question: Does the paper describe safeguards that have been put in place for responsible release of data or models that have a high risk for misuse (e.g., pretrained language models, image generators, or scraped datasets)?
    \item[] Answer: \answerNA{} 
    \item[] Justification: The paper poses no such risks

\item {\bf Licenses for existing assets}
    \item[] Question: Are the creators or original owners of assets (e.g., code, data, models), used in the paper, properly credited and are the license and terms of use explicitly mentioned and properly respected?
    \item[] Answer: \answerYes{} 
    \item[] Justification: We cited the related work,

\item {\bf New assets}
    \item[] Question: Are new assets introduced in the paper well documented and is the documentation provided alongside the assets?
    \item[] Answer: \answerYes{} 
    \item[] Justification: Our code have the README.md part.

\item {\bf Crowdsourcing and research with human subjects}
    \item[] Question: For crowdsourcing experiments and research with human subjects, does the paper include the full text of instructions given to participants and screenshots, if applicable, as well as details about compensation (if any)? 
    \item[] Answer: \answerNA{} 
    \item[] Justification: The paper does not involve crowdsourcing nor research with human subjects.

\item {\bf Institutional review board (IRB) approvals or equivalent for research with human subjects}
    \item[] Question: Does the paper describe potential risks incurred by study participants, whether such risks were disclosed to the subjects, and whether Institutional Review Board (IRB) approvals (or an equivalent approval/review based on the requirements of your country or institution) were obtained?
    \item[] Answer: \answerNA{} 
    \item[] Justification: The paper does not involve crowdsourcing nor research with human subjects.

\item {\bf Declaration of LLM usage}
    \item[] Question: Does the paper describe the usage of LLMs if it is an important, original, or non-standard component of the core methods in this research? Note that if the LLM is used only for writing, editing, or formatting purposes and does not impact the core methodology, scientific rigorousness, or originality of the research, declaration is not required.
    \item[] Answer: \answerYes{} 
    \item[] Justification: We described the usage of LLM.

\end{enumerate}